\documentclass{article}

\usepackage[preprint]{neurips_2025}

\usepackage[utf8]{inputenc} 
\usepackage[T1]{fontenc}    
\usepackage{hyperref}       
\usepackage{url}            
\usepackage{booktabs}       
\usepackage{amsfonts}       
\usepackage{nicefrac}       
\usepackage{microtype}      
\usepackage{array,multirow,booktabs}
\usepackage{caption}
\usepackage[dvipsnames,table,xcdraw]{xcolor}
\usepackage{mathrsfs}
\usepackage{graphics,latexsym,epsfig,psfrag,wrapfig,comment,paralist}
\usepackage{xspace}
\usepackage{mathdots}
\usepackage{bbold}
\usepackage{centernot}
\usepackage{algorithm}
\usepackage{algorithmic}
\usepackage{prettyref}
\usepackage{listings}
\usepackage{tikz}
\usepackage{pgfplots}
\usepackage{mdwmath}
\usepackage{mdwtab}
\usepackage{eqparbox}
\usepackage{multicol}
\usepackage{bigstrut,threeparttable}
\usepackage{bbm}
\usepackage{epstopdf}
\usepackage{graphicx}
\usepackage{setspace}
\usepackage{subcaption, bm}
\title{How Do LLMs Perform Two-Hop Reasoning in Context?}

\author{%
  Tianyu Guo$^{1}\thanks{Equal contribution.}$ \quad
  Hanlin Zhu$^{2}\footnotemark[1]$ \quad
  Ruiqi Zhang$^1$ \quad \\
  \textbf{Jiantao Jiao$^2$ \quad
  Song Mei$^{1,2}$ \quad
  Michael I. Jordan$^{1,2}$ \quad
  Stuart Russell$^2$} \\
  \\
  $^1$Department of Statistics, UC Berkeley \\
  $^2$Department of EECS, UC Berkeley \\
  \\
  \texttt{\{tianyu\_guo, hanlinzhu\}@berkeley.edu}
}

\def\shownotes{0}  
\ifnum\shownotes=1
\newcommand{\authnote}[2]{{\scriptsize $\ll$\textsf{#1 notes: #2}$\gg$}}
\else
\newcommand{\authnote}[2]{}
\fi
\newcommand\RZ[1]{\textcolor{red}{\authnote{RZ}{#1}}}
\newcommand\hanlin[1]{\textcolor{blue}{\authnote{Hanlin}{#1}}}

\newcommand\tianyu[1]{\textcolor{brown}{\authnote{Tianyu}{#1}}}
\usepackage[textsize=tiny]{todonotes}

\usepackage{amsmath}
\usepackage{amssymb}
\usepackage{mathtools}
\usepackage{amsthm}
\usepackage[dvipsnames,table]{xcolor}
\usepackage[most]{tcolorbox}

\definecolor{linkColor}{HTML}{E74C3C}
\definecolor{pearcomp}{HTML}{B97E29}
\definecolor{citeColor}{HTML}{2980B9}
\definecolor{urlColor}{HTML}{1D2DEC}
\definecolor{conjColor}{HTML}{9ab569}
\usetikzlibrary{shadows}
\newtheorem{hypothesis}{Hypothesis}[section]
\newcounter{exa}
\renewcommand{\theexa}{\arabic{exa}} 
\newtheoremstyle{break}
  {\topsep}{\topsep}%
  {\itshape}{}%
  {\bfseries}{}%
  {\newline}{}%
\tcbset{
  myexample/.style={
    enhanced,
    colback=yellow!10!white,
    colframe=red!50!black,
    fonttitle=\scshape,
    titlerule=0pt,
    attach boxed title to top center={yshift=-2mm},
    boxed title style={colback=yellow!10!white, colframe=red!50!black},
    coltitle=red!50!black,
    drop shadow,
    title={\refstepcounter{exa}Example~\theexa:\quad #1},
    highlight math style={reset,colback=LightBlue!50!white,colframe=Navy}
  }
}

\newtcolorbox{texample}[1][]{myexample={#1}}

\usepackage[capitalize,noabbrev]{cleveref}

\newcommand{\attlogit}{\texttt{Attn-Logit}}
\newcommand{\logit}{\texttt{Logit}}
\newcommand{\readout}{\texttt{ReadOut}}

\newcommand{\brg}{\textsc{Brg}}

\newcommand{\child}{\textsc{Child}}
\newcommand{\parent}{\textsc{Parent}}

\newcommand{\Src}[1][]{\if\relax\detokenize{#1}\relax$[\textsc{Src}]$\else$[\text{Src}_{#1}]$\fi}
\newcommand{\Brg}[1][]{\if\relax\detokenize{#1}\relax$[\textsc{Brg}]$\else$[\text{Brg}_{#1}]$\fi}
\newcommand{\Ed}[1][]{\if\relax\detokenize{#1}\relax$[\textsc{End}]$\else$[\text{End}_{#1}]$\fi}
\newcommand{\targetsource}[1][]{\if\relax\detokenize{#1}\relax$[\textsc{Src-T}]$\else$[\text{Src-T}_{#1}]$\fi}
\newcommand{\targetbridge}[1][]{\if\relax\detokenize{#1}\relax$[\textsc{Brg-T}]$\else$[\text{Brg-T}_{#1}]$\fi}
\newcommand{\targetend}[1][]{\if\relax\detokenize{#1}\relax$[\textsc{End-T}]$\else$[\text{End-T}_{#1}]$\fi}
\newcommand{\nontargetsource}[1][]{\if\relax\detokenize{#1}\relax$[\textsc{Src-NT}]$\else$[\text{Src-NT}_{#1}]$\fi}
\newcommand{\nontargetbridge}[1][]{\if\relax\detokenize{#1}\relax$[\textsc{Brg-NT}]$\else$[\text{Brg-NT}_{#1}]$\fi}
\newcommand{\nontargetend}[1][]{\if\relax\detokenize{#1}\relax$[\textsc{End-NT}]$\else$[\text{End-NT}_{#1}]$\fi}
\newcommand{\Token}[1][]{\if\relax\detokenize{#1}\relax$[\textsc{Token}]$\else$[\text{Token}_{#1}]$\fi}
\newcommand{\Qry}[1][]{\if\relax\detokenize{#1}\relax$\textsc{Qry}$\else$\textsc{Qry}({#1})$\fi}
\newcommand{\Key}[1][]{\if\relax\detokenize{#1}\relax$\textsc{Key}$\else$\textsc{Key}({#1})$\fi}
\newcommand{\Val}[1][]{\if\relax\detokenize{#1}\relax$\textsc{Val}$\else$\textsc{Val}({#1})$\fi}
\newcommand{\AttentionWeight}[1][]{\if\relax\detokenize{#1}\relax$\textsc{Attn-Weight}$\else$\textsc{Attn-Weight}({#1})$\fi}
\newcommand{\Softmax}[1][]{\if\relax\detokenize{#1}\relax$\textsc{Softmax}$\else$\textsc{Softmax}({#1})$\fi}

\newcommand{\Bos}{\textsc{<BOS>}}
\newcommand{\Readout}{\textsc{ReadOut}}

\newcommand{\pseudosoftmax}{\widetilde{\mathcal{S}}}

\newcommand{\buffer}{\textsc{buf}}
\newcommand{\content}{\textsc{cont}}
\newcommand{\R}{\mathbb{R}}

\begin{document}

\maketitle

\footnotetext{Our code is available at  \href{https://github.com/GuoTianYu2000/twohopIC}{https://github.com/GuoTianYu2000/twohopIC}.}
\begin{abstract}
``Socrates is human. All humans are mortal. Therefore, Socrates is mortal.'' 
This form of argument illustrates a typical pattern of two-hop reasoning.
Formally, two-hop reasoning refers to the process of inferring a conclusion by making two logical steps, each connecting adjacent concepts, such that the final conclusion depends on the integration of both steps.
It is one of the most fundamental components of human reasoning and plays a crucial role in both formal logic and everyday decision-making.
Despite recent progress in large language models (LLMs), we surprisingly find that they can fail at solving simple two-hop reasoning problems when distractors are present. 
we observe on a synthetic dataset that pre-trained LLMs often resort to random guessing among all plausible conclusions. 
However, after few steps of fine-tuning, models achieve near-perfect accuracy and exhibit strong length generalization. 
To understand the underlying mechanisms, we train a 3-layer Transformer from scratch on a synthetic two-hop reasoning task and reverse-engineer its internal information flow.
We observe a clear progression in the attention logits throughout training.
This pictures a sharp phase transition from an initial stage of random guessing to the emergence of a structured sequential query mechanism, where the model first retrieves the preceding and the bridge concepts in the early layers and then uses them to infer the final answer. 
Finally, we show that these dynamics can be captured by a minimal three-parameter attention-only network.
\end{abstract}

\section{Introduction}\label{sec:intro}

Modern Large Language Models (LLMs) trained on language data have shown impressive abilities in solving complex reasoning tasks, including linguistic, mathematical, and programming problems \citep{cobbe2021training,wei2022chain,achiam2023gpt,dubey2024llama,yang2024qwen2,shao2024deepseekmath,guo2025deepseek}.
Among these, multi-hop reasoning stands out as one of the most fundamental and important forms of reasoning.
It refers to the process of drawing conclusions by integrating information across multiple intermediate steps or pieces of evidence.
For example, consider the following chain of facts: \textit{Marie Curie was a physicist, physicists study matter, and matter is fundamental to understanding the universe}. From these premises, we can infer that \textit{Marie Curie contributed to understanding the universe.}
This reasoning requires multiple inferential steps, each built on the previous one, making the reasoning process \textit{inherently compositional}. 
Such multi-hop reasoning is critical for tasks where information must be integrated across several logical or semantic connections. 
Despite its simplicity and clear structure, it still lacks a clear understanding of \textit{whether and how} pre-trained LLMs can \textit{reliably} perform such reasoning.

In this paper, we study the mechanism behind \textbf{two-hop reasoning}, which is the simplest and most tractable form of multi-hop reasoning.
We define a premise as a factual statement that connects two entities through a specific relation.
In two-hop reasoning, two such premises form a chain that allows one to infer a final conclusion.
We define the source entity (\Src) as the starting point, the bridge entity (\Brg) as the intermediate link, and the end entity (\text{\Ed}) as the target of inference.
Here is a concrete example that illustrates the above reasoning procedure:
\[
\begin{aligned}
(\text{Premises})\quad
&\underbrace{\text{Socrates}}_{\hbox{\Src}}%
\;\text{is}\;
\underbrace{\text{human}}_{\hbox{\Brg}}\,.%
\quad
\underbrace{\text{Humans}}_{\hbox{\Brg}}%
\;\text{are}\;
\underbrace{\text{mortal}}_{\hbox{\text{\Ed}}}\,.\\
(\text{Conclusion})\quad
&
\underbrace{\text{Socrates}}_{\hbox{\Src}}%
\;\text{is}\;
\underbrace{\text{mortal}}_{\hbox{\text{\Ed}}}\,.
\end{aligned}
\]

We begin by building a test dataset of simple two-hop reasoning examples with distractors, where multiple unrelated two-hop examples are presented in the same context.
We observe that pretrained LLMs struggle with it.
In particular, the models tend to \textit{guess randomly}, with accuracy dropping to about $1/K$ (where $K$ is the number of two-hop examples).
This suggests that distractors in the context strongly confuse the model and hurt the performance, revealing a weakness in its reasoning ability.
After fine-tuning the model on a curated two-hop dataset, we observe a sharp transition in performance—from random guessing to near-perfect accuracy.
Even more impressively, the fine-tuned model generalizes to harder settings with more distractors.

To better understand this behavior, we study a three-layer Transformer.
When trained on a curated symbolic two-hop reasoning dataset with distractors, the model exhibits a sharp phase transition moving from \textit{uniform guessing} to a \textit{structured reasoning phase}, where it consistently performs correct inference.
By fully reverse-engineering the model, we analyze how information flows across layers and identify the specific role each layer plays.
We show that this transition is driven by particular patterns in the attention logits, which guide the model to retrieve relevant entities in the correct order.
In the structured reasoning phase, the model performs two-hop inference by sequentially attending to the source and then the bridge entity to reach the correct end entity.
To further support our findings, we also built a simple three-parameter analytical model that captures the full dynamics of the three-layer Transformer.

Our contributions and paper outline are summarized as follows.
\begin{itemize}
    \item We introduce \textit{two-hop reasoning with distractors}, and show pretrained LLMs collapse to random guessing.
    Fine-tuning leads to a sharp improvement and a strong generalization performance (\Cref{sec:llm-twohop}).
    \item We reverse-engineer a three-layer Transformer and restore the transition from random guessing to the emergence of a sequential query mechanism. We reveal how each layer contributes to the emergence of reasoning ability through structured attention patterns (\Cref{sec:tf-twohop}).
\end{itemize}

\subsection{Related works}
\label{subsec:related_work}

In this part, we review and discuss additional related papers.

\paragraph{Multi-hop reasoning.}
Our research focuses on multi-hop reasoning, a fundamental form of reasoning and a key benchmark for evaluating LLMs \citep{zhong2023mquake}. 
Recent studies such as \citet{yang2024latent} have shown that LLMs can follow latent reasoning paths when given certain types of prompts.
A growing body of papers focused on mechanistically understanding how LLMs perform multi-hop reasoning by sequentially attending to intermediate steps \citep{biran2024hopping, wang2024grokked, feng2024extractive}.
Prior research mainly focused on the \textit{in-weight} multi-hop reasoning, where the model retrieves and combines factual knowledge stored in its internal weights.
In contrast, our work focused on \textit{in-context} two-hop reasoning, where the model must extract relevant facts directly from the context and reason on-the-fly without relying on memorized knowledge. \RZ{I think there may be more papers about two-hop reasoning?} \tianyu{to-do}

\paragraph{In-context learning and induction head.}
Transformer-based models exhibit strong In-Context Learning (ICL) ability.
It refers to the ability to predict the label of new input simply from several demonstrations, without updating model weights \citep{brown2020language}.
Transformers have been shown to solve various tasks in context, such as regression, classification \citep{akyurek2022learning, garg2022can, von2023transformers, zhang2024trained, ahn2024transformers, huang2023context, nichani2024transformers}, Bayesian inference \citep{xie2021explanation}, model selection \citep{bai2023transformers}, and sequential decision making \citep{lin2023transformers}.
Our study focuses on two-hop reasoning in context, which is very different and more complex than the standard regression-style ICL tasks.

Moreover, prior work showed that ICL relies on the induction head--a pattern of attention that enables the model to copy and complete sequences by linking repeated tokens \citep{elhage2021mathematical,olsson2022context}.
Several recent theoretical and empirical analyses have extensively studied induction-head mechanisms in small transformers \citep{bietti2023birth, nichani2024transformers, wang2024transformers, chen2024unveiling}, showing that a two-layer transformer is required to perform induction-head tasks \citep{sanford2024one}. 
In comparison, two-hop reasoning is a more complex extension to the induction head and requires the model to combine two separate relational facts.
We show that two-layer Transformers cannot solve this task; the minimal architecture required is a three-layer Transformer.
Theoretically, it has been shown that a Transformer with $\log k$ layers is both necessary and sufficient to perform $k$-hop reasoning in context \citep{sanford2024transformers}.

\paragraph{Interpretability of LLMs.}
Beyond ICL and the induction head, many studies have aimed to interpret the internal mechanisms of LLMs \citep{charton2022my, liu2022towards, allen2023physics, zhu2023physics, guo2023transformers, zhang2022unveiling}.
This includes works on grokking \citep{nanda2023progress}, function vectors \citep{todd2023function}, circuit discovery \citep{elhage2021mathematical,wang2022interpretability,conmy2023towards,shi2024hypothesis,hase2024does}, 
the binding ID mechanism \citep{feng2023language}, and the association-storage mechanism \citep{meng2022locating,geva2023dissecting}. 
Our work is not directly comparable with theirs.

Methodologically, a growing body of studies has focused on designing small systems, where essentially the same phenomenon can be observed, and then dissecting the proxy model to interpret the mechanism of LLMs.
For example, \citet{olsson2022context} replicates the induction head in a minimal two-layer network.
\citet{bietti2024birth} further explains the rapid emergence of bigram memorization and the slower development of an induction head. 
\citet{zhu2024towards} theoretically analyzed the cause of the reversal curse in bilinear models and one-layer transformers.
\citet{reddy2023mechanistic} studies the abrupt emergence of induction heads in two-layer models and captured the underlying mechanism using a two-parameter toy model.
\citet{guo2024active} reproduces the extreme-token phenomenon in Transformers with one to three layers on the Bigram-Backcopy task, and then identifies the mechanism of active and dormant attention head in small and large models.
Our work is methodologically similar to this line of work, whereas two-hop reasoning is a more complex tasks than Bigram-Backcopy or copy-paste, so the revealed mechanism is much more complex.

\section{Two-Hop Reasoning in LLMs}
\label{sec:llm-twohop}
\subsection{Task and data}
\paragraph{Two-hop reasoning.}
We begin by formally defining the task of two-hop reasoning.
Each reasoning chain involves three distinct entities:

\begin{itemize}
\item \textbf{Source entity \Src{}:} The initial entity from which reasoning originates.
\item \textbf{Bridge entity \Brg{}:} An intermediate entity that connects the source entity to the final inferred entity.
\item \textbf{End entity \text{\Ed}{}:} The target entity that the reasoning aims to infer.
\end{itemize}

A valid two-hop reasoning task consists of exactly two premises. The first premise connects the source entity \Src{} to the bridge entity \Brg{}, and the second premise connects the bridge entity \Brg{} to the end entity \text{\Ed}{}. Together, these premises form a logical chain that supports drawing a conclusion from the source to the end entity.

\paragraph{Two-hop reasoning with distractors.}
To evaluate LLMs' robustness in reasoning, we introduce \textit{two-hop reasoning with distractors}.
This setting intentionally incorporates irrelevant premises to assess the robustness and precision of LLM reasoning.
In this setting, multiple two-hop reasoning chains are provided within the same context, but only one chain (the \textit{target chain}) leads to the correct inference. The other chains, meanwhile, serve purely as \textit{distractors}, introducing irrelevant entities.
We aim to infer the end entity of the target chain. The entities involved in the reasoning chains are categorized into \textit{target entities} and \textit{non-target entities}. 
Target entities, denoted as \targetsource, \targetbridge, and \targetend, correspond respectively to the source, bridge, and end entities within the target reasoning chain. 
Non-target entities, denoted as \nontargetsource, \nontargetbridge, and \nontargetend, represent the entities involved in distractor chains, which do not contribute to the correct inference.

\paragraph{Dataset.}
To systematically evaluate two-hop reasoning performance, we generate a synthetic dataset based on standardized logical templates. Templates represent structured reasoning patterns from domains such as geography, biology, and arithmetic. 
Templates follow a consistent structure, such as "[A] is the father of [B]. [B] is the father of [C]. Therefore, [A] is the grandfather of [C]." 
Here, [A], [B], and [C] represent source, bridge, and end entities, respectively.

For each data sample, we randomly select a single template and generate multiple argument chains by populating placeholders with entity sets sampled from a predefined entity pool. 
Exactly one chain is designated as the target reasoning chain, and its corresponding conclusion part is presented at the end of the context as a query. All remaining chains act as distractors. Only two premises of the distractors are present in the context.
Target and non-target reasoning premises in the context are randomly permuted.
We construct a dataset comprising more than 50,000 such reasoning contexts, spanning 6 distinct templates. This rigorous dataset construction provides a robust evaluation framework to investigate how effectively LLMs distinguish relevant from irrelevant premises and accurately perform two-hop logical reasoning.
A concrete example of such a two-hop reasoning chain with distractors is shown in the following example.

\begin{texample}[An Example of Two-hop Reasoning with 2 Distractors]\label{example.two.hop.with.distractor}
Question: \textcolor{red}{John} is the father of \textcolor{red}{Paul}. 
\textcolor{blue}{Luke} is the father of \textcolor{blue}{Tom}.
\textcolor{blue}{Sam} is the father of \textcolor{blue}{Joe}.
\textcolor{red}{Paul} is the father of \textcolor{red}{Ben}.
\textcolor{blue}{Tom} is the father of \textcolor{blue}{Mark}.
\textcolor{blue}{Joe} is the father of \textcolor{blue}{Max}.
Therefore, \textcolor{red}{John} is the grandfather of \textcolor{red}{???}\\
Answer: \textcolor{red}{Ben}.\\[5pt]

\textcolor{red}{Red}: Target source/bridge/end entities in the target chain.\\
\textcolor{blue}{Blue}: Non-target source/bridge/end entities in the non-target chain.\\[5pt]

Probability assignment:\\
Base model: \{`\textcolor{red}{Ben}':0.33, `\textcolor{blue}{Mark}': 0.32, `\textcolor{blue}{Max}': 0.31,...\}.\\
Fine-tuned model: \{`\textcolor{red}{Ben}':0.97, `\textcolor{blue}{Mark}': 0.01, `\textcolor{blue}{Max}': 001,...\}.
\end{texample}

\subsection{Results}
\paragraph{Pre-trained LLMs perform random guessing at the presence of distractors.}
We evaluate the performance of the Llama2-7b-base model on two-hop reasoning tasks both with and without distractors. 
Specifically, on a held-out test set, we track the probability assigned by the model to the first token of the target end entity (\targetend{}) versus other tokens at the conclusion of each context.
Surprisingly, the model achieves high accuracy in identifying the \targetend{} when no distractors are present, but exhibits a dramatic drop in performance even with a single distractor. 
As the number of distractors increases, accuracy further decreases. 
In fact, \Cref{tab.random.guessing} shows the next-token probability assigned to the (first token of) every possible end entities in the context (\targetend{} and all \nontargetend{} entities) are approximately $1/K$ (although there is a slight bias toward \targetend{}), where $K$ is the total number of reasoning chains presented. 
Therefore, the model almost resorts to \textit{random guessing} among the set of possible end entities.
This behavior is clearly illustrated by Example 1 and \Cref{tab.random.guessing} and highlights the vulnerability of pre-trained models to distractors in two-hop reasoning tasks.

\paragraph{Fine-tuned LLMs significantly improve accuracy and generalization.}
To address the challenge posed by distractors, we fine-tune the Llama2-7b model using 1,000 curated prompts, each containing exactly one target reasoning chain and one distractor chain (further details provided in~\Cref{sec:llm}). We then evaluate the fine-tuned model on the same held-out test set. \Cref{tab.random.guessing} compares model performance before and after fine-tuning across contexts with varying numbers of distractors.

Before fine-tuning, the Llama2-7b model correctly predicts the \targetend{} only in distraction-free contexts and defaults to random guessing when distractors are introduced. However, after fine-tuning, the model reliably identifies the \targetend{} even in the presence of multiple distractors. Remarkably, despite being fine-tuned exclusively on contexts with a single distractor, the model generalizes effectively to scenarios containing multiple distractors, accurately performing two-hop reasoning tasks with as many as five distractor chains. These findings confirm that fine-tuning significantly enhances LLM robustness and generalization capabilities.

\vspace{2em}
\begin{table}[tb]
\begin{center}
\caption{\small{
Performance comparison of Llama2-7b base and fine-tuned models on two-hop reasoning tasks with varying numbers of distractors. 
$K=1$ indicates the scenario without distractors. 
The \targetend{} and \nontargetend{} rows report the average probabilities assigned to the first token of the target end entity and non-target end entities, respectively. 
For cases with multiple distractors, the reported probabilities for non-target entities are averaged first over all \nontargetend{} within each context, then across all contexts. 
The values in the parentheses are the standard errors.
} 
}
\vspace{1mm}
\resizebox{1\textwidth}{!}{
\begin{tabular}{c|c|ccccc}
\toprule[1pt]
\midrule
\multicolumn{1}{c|}{\multirow{2}{*}{\textbf{Models}}} &
\multicolumn{1}{c|}{\multirow{2}{*}{\textbf{Next Tokens}}} & 
\multicolumn{5}{c}{\textbf{The Number of Reasoning Chains in the Context $(K)$}} \\
\cmidrule{3-7}
& & \begin{tabular}{c} 1 \end{tabular}
& \begin{tabular}{c} 2 \end{tabular}
& \begin{tabular}{c} 3 \end{tabular}
& \begin{tabular}{c} 4 \end{tabular}
& \begin{tabular}{c} 5 \end{tabular}\\ 
\midrule
\multicolumn{1}{c|}{\multirow{2}{*}{\textbf{Llama2-7b}}} & \targetend & 0.82 (0.01) & 0.41 (0.01) & 0.31 (0.01) & 0.25 (0.01) & 0.20 (0.01) \\
 & \nontargetend & NA (NA) & 0.34 (0.01) & 0.20 (0.01) & 0.14 (0.00) & 0.11 (0.00) \\
\midrule
\multicolumn{1}{c|}{\multirow{2}{*}{\shortstack{\textbf{Fine-tuned}\\ \textbf{on $K=2$}}}} & \targetend & 1.00 (0.00) & 1.00 (0.00) & 0.92 (0.01) & 0.89 (0.01) & 0.86 (0.01) \\
 & \nontargetend & NA (NA) & 0.00 (0.00) & 0.04 (0.01) & 0.03 (0.01) & 0.03 (0.00) \\
\midrule
\bottomrule
\end{tabular}
}
\label{tab.random.guessing}
\end{center}
\end{table}

\section{Two-Hop Reasoning in Three-Layer Transformers}
\label{sec:tf-twohop}
\subsection{Data and models}

\paragraph{Symbolic two-hop reasoning task.}
To systematically investigate the mechanisms underlying random guessing and how Transformers learn two-hop reasoning from context, we design a symbolic version of the two-hop reasoning task. 
In this simplified setting, we hide the predicates of reasoning chains and represent each entity with a unique single token.
With a little abuse of notation, these tokens are denoted as the \textit{source token} (\Src{}), \textit{bridge token} (\Brg{}), and \textit{end token} (\text{\Ed}{}), respectively.
Each symbolic two-hop reasoning chain consists of two premises represented by concatenated token sequences: the first premise is \Src{} \Brg{}, and the second premise is \Brg{} \text{\Ed}{}. 
The conclusion part of each chain is represented by the token sequence \Src{} \text{\Ed}{}. 
For each premise, we call the paired tokens the \textit{parent token} and the \textit{child token}.
For example, in the first premise \Src{} \Brg{}, we call \Src{} the parent token and \Brg{} the child token.
Below is an example of the reasoning chain from our symbolic task.
\begin{align*}
    \underbrace{\text{\Src} \hspace{1em} \text{\Brg}}_{\text{The first premise}}
    \hspace{1em}
    \underbrace{\text{\Brg} \hspace{1em} \text{\text{\Ed}}}_{\text{The second premise}}
    \hspace{1em}
    \underbrace{\text{\Src} \hspace{1em} \text{\text{\Ed}}}_{\text{The conclusion}}
\end{align*}

For each context, we randomly sample five unique two-hop reasoning chains, each with distinct source, bridge, and end tokens. 
One chain is randomly chosen as the \textit{target reasoning chain}, while the remaining four serve as \textit{distractors}. 
Tokens within the target reasoning chain are denoted as \targetsource{}, \targetbridge{}, and \targetend{}, and tokens within distractor chains as \nontargetsource{}, \nontargetbridge{}, and \nontargetend{}.
Each context includes all premises from these five reasoning chains and concludes with the source token of the target chain as a \textit{query}. 
Premises are randomly permuted within the context, but the order within each reasoning chain remains fixed, that is, the source-to-bridge premise always precedes the bridge-to-end premise in the context.
An illustrative example from our simulated dataset is presented below. 
Note that all contexts in our symbolic dataset have the same length.

\begin{texample}[Symbolic Two-Hop Reasoning with Four Distractors]\label{example.symbolic}
Context: \Bos{} \textcolor{blue}{\nontargetsource[1]} \ \textcolor{blue}{\nontargetbridge[1]} \ \textcolor{blue}{\nontargetsource[2]} \ \textcolor{blue}{\nontargetbridge[2]} \ \textcolor{red}{\targetsource} \ \textcolor{red}{\targetbridge} \ \textcolor{blue}{\nontargetsource[3]} \ \textcolor{blue}{\nontargetbridge[3]} \ \textcolor{blue}{\nontargetsource[4]} \ \textcolor{blue}{\nontargetbridge[4]} \
\textcolor{blue}{\nontargetbridge[3]} \ \textcolor{blue}{\nontargetend[3]} \ \textcolor{red}{\targetbridge} \ \textcolor{red}{\targetend} \ \textcolor{blue}{\nontargetbridge[1]} \ \textcolor{blue}{\nontargetend[1]} \ \textcolor{blue}{\nontargetbridge[4]} \ \textcolor{blue}{\nontargetend[4]} \ \textcolor{blue}{\nontargetbridge[2]} \ \textcolor{blue}{\nontargetend[2]} \ \targetsource ?\\[5pt]
\Bos{}: The begin-of-sequence token.\\
\textcolor{red}{Red}: Tokens from the target chain.\\
\textcolor{blue}{Blue}: Tokens from distractor chains. Subscripts distinguish different reasoning chains.\\
\targetsource: The query token.
\end{texample}

\begin{figure}[htbp]
    \centering
    \begin{subfigure}{0.4\textwidth}
        \centering
\caption{ \small{Training loss}}
\label{fig:log_acc_dynamics}
\includegraphics[width=\textwidth]{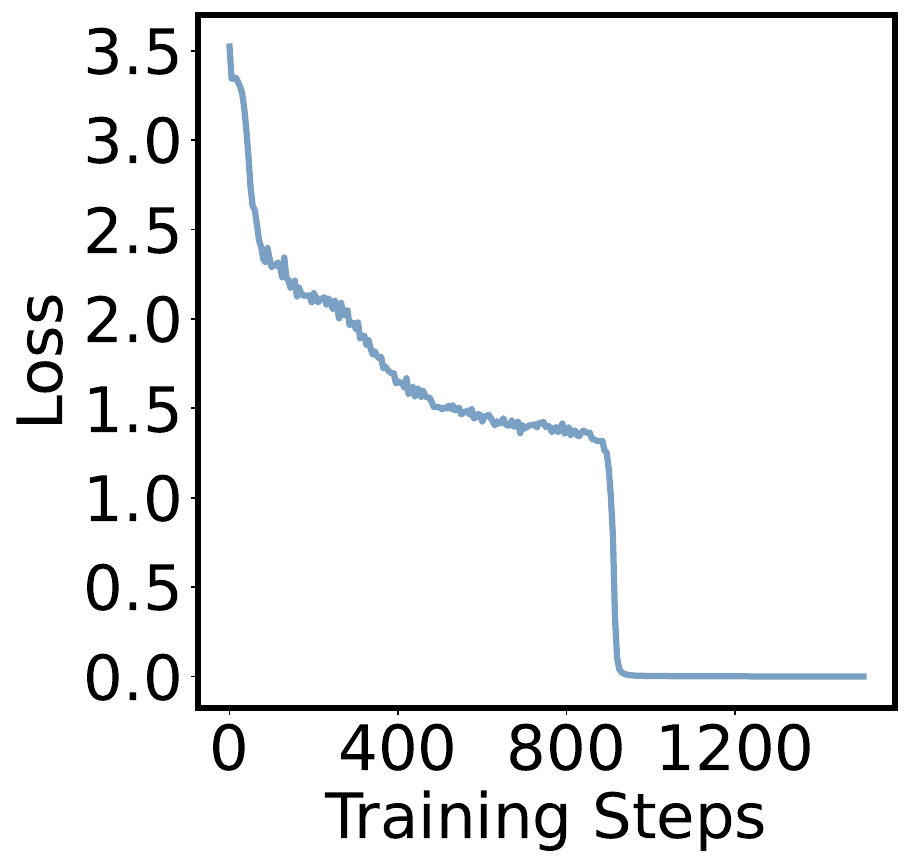}
    \end{subfigure}
    \hfill
    \begin{subfigure}{0.4\textwidth}
        \centering
\caption{ \small{Predicted probabilities}}
\label{fig:acc_dynamics}
\includegraphics[width=\textwidth]{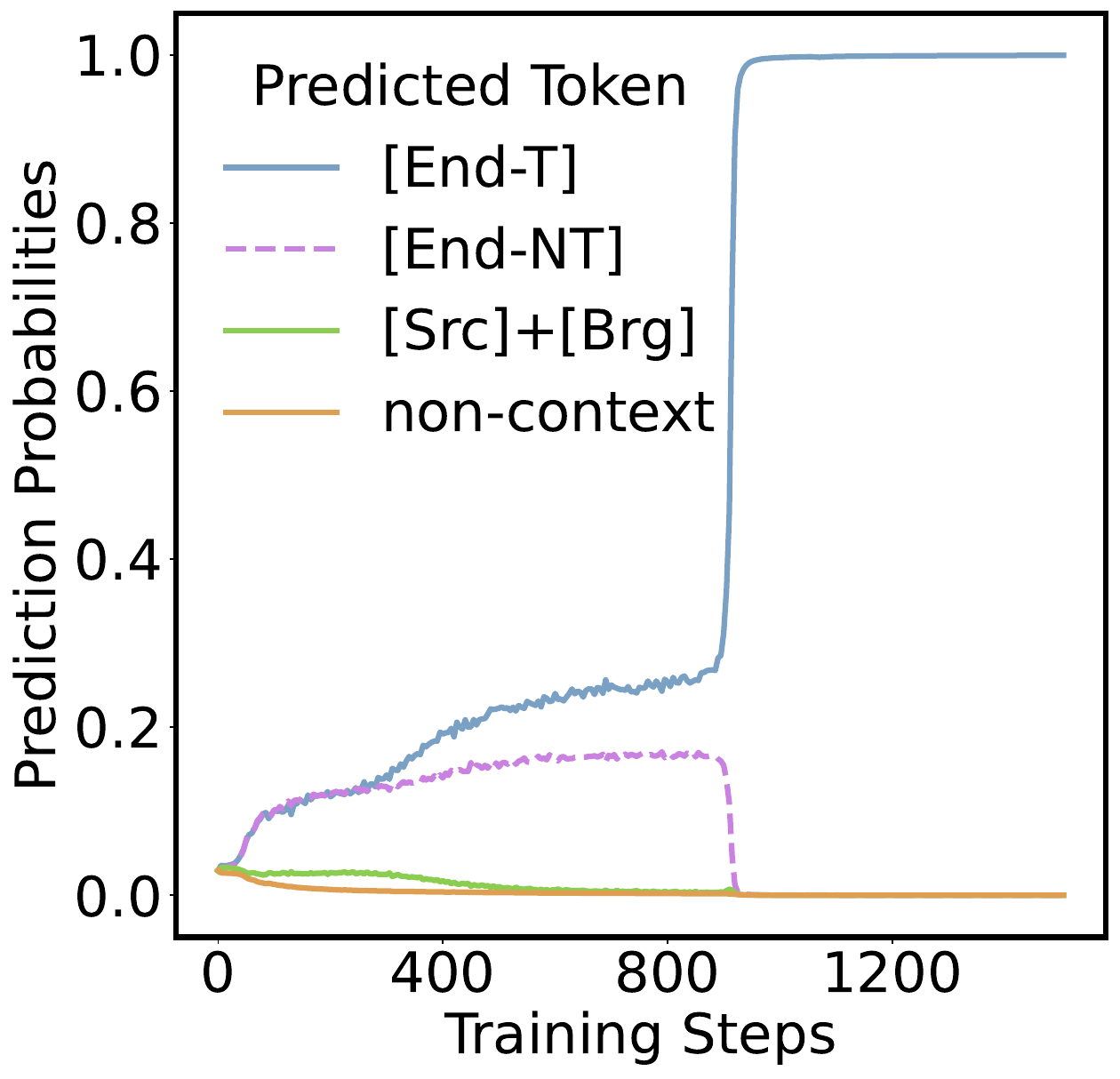}
    \end{subfigure}
\caption{\textbf{The loss and the predicted probabilities.} \textit{Left (a):} The cross entropy loss computed at the query token, with the label being the correct target end token (\targetend{}) in the preceding premises. \textit{Right (b):} The predicted probabilities for different tokens throughout training. The \nontargetend{} line represents probabilities averaged across \nontargetend[1], \nontargetend[2], \nontargetend[3], and \nontargetbridge[4]. Before approximately 
800 steps, the \nontargetend{} and \targetend{} lines remain close, indicating an almost random guessing behavior during this \textit{slow learning phase}.}
\end{figure}

\paragraph{Three-layer Transformer Analysis.}
We investigate the minimal Transformer architecture capable of capturing both the random guessing phenomenon and the structured learning phase observed in two-hop reasoning tasks. 
By comparing Transformers of varying depths, we find that a three-layer Transformer with a single attention head per layer is the minimal structure required. 
Figures~\ref{fig:log_acc_dynamics} and~\ref{fig:acc_dynamics} illustrate that when trained on our symbolic two-hop reasoning dataset, the three-layer Transformer exhibits a sharp phase transition at approximately 800 training steps. 
Before this transition, the model assigns nearly uniform probabilities (around 0.2) to all possible end tokens (\targetend{} and all \nontargetend{} tokens), albeit with a slight bias toward \targetend{}.
This indicates an almost random guessing behavior in this \textit{slow learning phase}.

In contrast, single-layer and two-layer Transformers fail to consistently solve the symbolic task, even after extensive training.
Additionally, we observe that removing the MLP layers from the three-layer Transformer (thus using an attention-only Transformer) does not harm the observed learning dynamics or the phase transition. 
Therefore, our subsequent analysis will focus on the three-layer attention-only Transformers.

\subsection{A Methodology for Reverse Engineering Transformers}
In the following sections, we reverse-engineer how a three-layer attention-only Transformer learns to perform two-hop reasoning by analyzing its internal information flow during training. 
We start by introducing two primary methodologies for empirical analysis: examining \textit{attention logits} and applying the \textit{logit lens} technique.

\paragraph{Attention logits.}
Our first method analyzes the Transformer's attention mechanism, which controls how information flows between tokens.
Specifically, we examine the \textit{attention logits}, the raw scalar values computed immediately before the softmax operation within the attention layer.
These logits quantify how strongly one token retrieves information from another. Plotting attention logits also reduces the complications induced by the softmax operation.
For clarity, we refer to the values after the softmax operation as \textit{attention weights}. 
We visualize attention logits across tokens and layers, highlighting entries with notably large values. 
When a token \textit{attends to} another, the corresponding attention logit is high, indicating the information retrieval. 
This retrieval process involves copying part or all of the information from the key token into a buffer, potentially separate from the query token’s own information storage.


\paragraph{Logit lens.}
Our second method employs the \textit{logit lens} approach \citep{belrose2023eliciting}, which interprets hidden states within the Transformer by projecting them directly onto the output space used for next-token prediction. 
In standard Transformers, the logits at the final layer are computed using a \Readout{} operator, which applies layer normalization followed by a linear transformation into the vocabulary space.
These logits, calculated at the final token position (the query token), are then converted into probabilities through softmax.

Formally, let $h_{\mathsf{query}}$ be the hidden state at the query token in the final layer, and \AttentionWeight[i] the attention weight between the $i$-th token and the query token. The probability of the next token is approximated by:
\begin{align*}
    \textsc{Softmax} (\textsc{ReadOut} (h_{\mathsf{query}}))
    \approx 
    \textsc{Softmax} \left(\sum_{i} \textsc{AttentionWeight} (i) \cdot \textsc{ReadOut} (\textsc{Val}(i))\right),
\end{align*}
where the summation includes all preceding tokens. 
In the following sections, we track these vectors after the readout operator in the earlier layers to clarify how the final logits and probabilities are computed.

\subsection{Mechanistic interpretation for the slow learning phase}

Phenomenologically, during the slow learning phase (prior to the sharp phase transition), the model effectively resorts to random guessing among all possible end tokens (\targetend{} and \nontargetend{} tokens). To explain this behavior, we closely examine attention logits across each Transformer layer. 

\begin{figure}[ht]
    \centering
    \begin{subfigure}{0.3\textwidth}
        \centering
        \caption{\small{Layer 1}}
        \label{fig:step800-layer-1}
        \includegraphics[width=\textwidth]{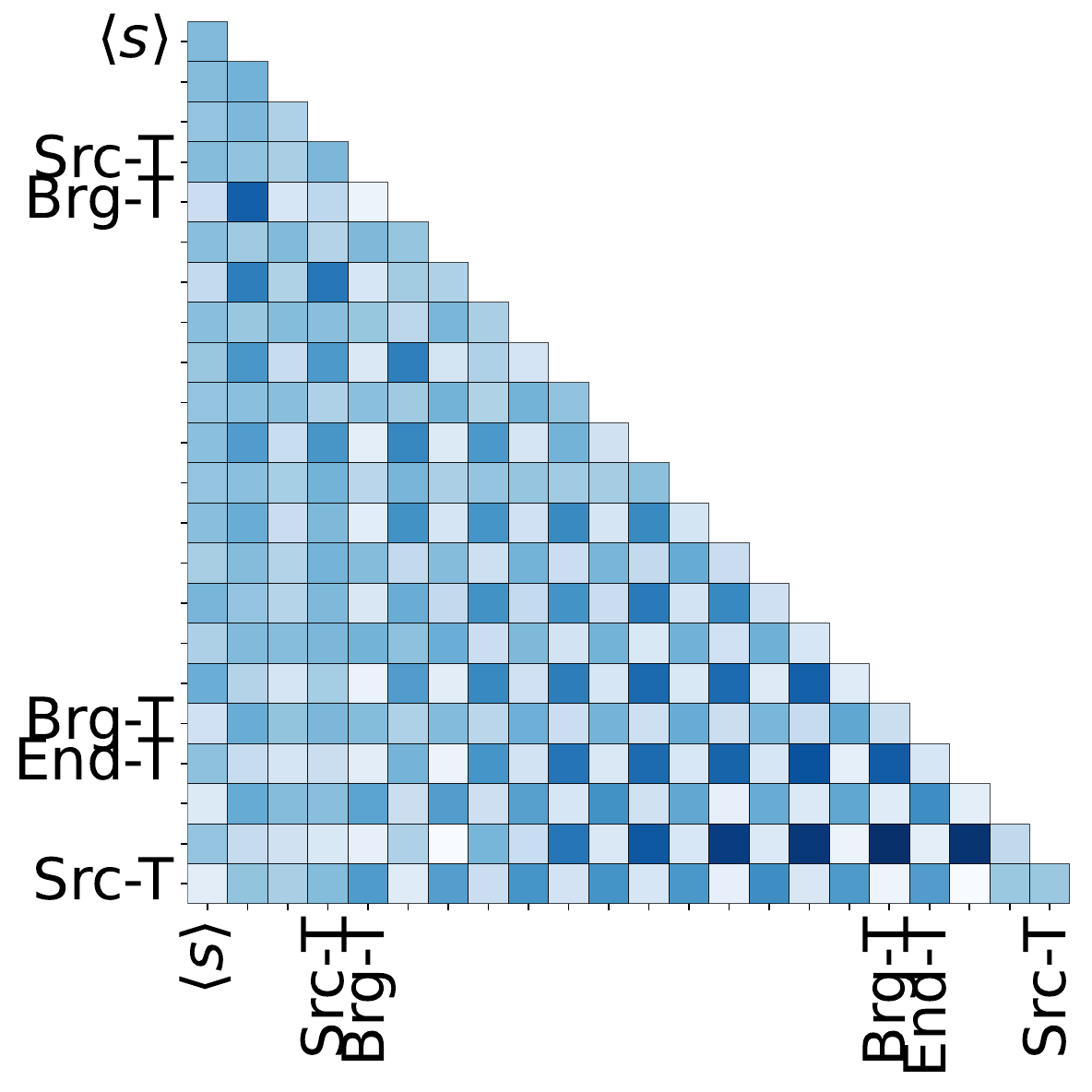}
    \end{subfigure}
    \hfill
    \begin{subfigure}{0.3\textwidth}
        \centering
        \caption{\small{Layer 2}}
        \label{fig:step800-layer-2}
        \includegraphics[width=\textwidth]{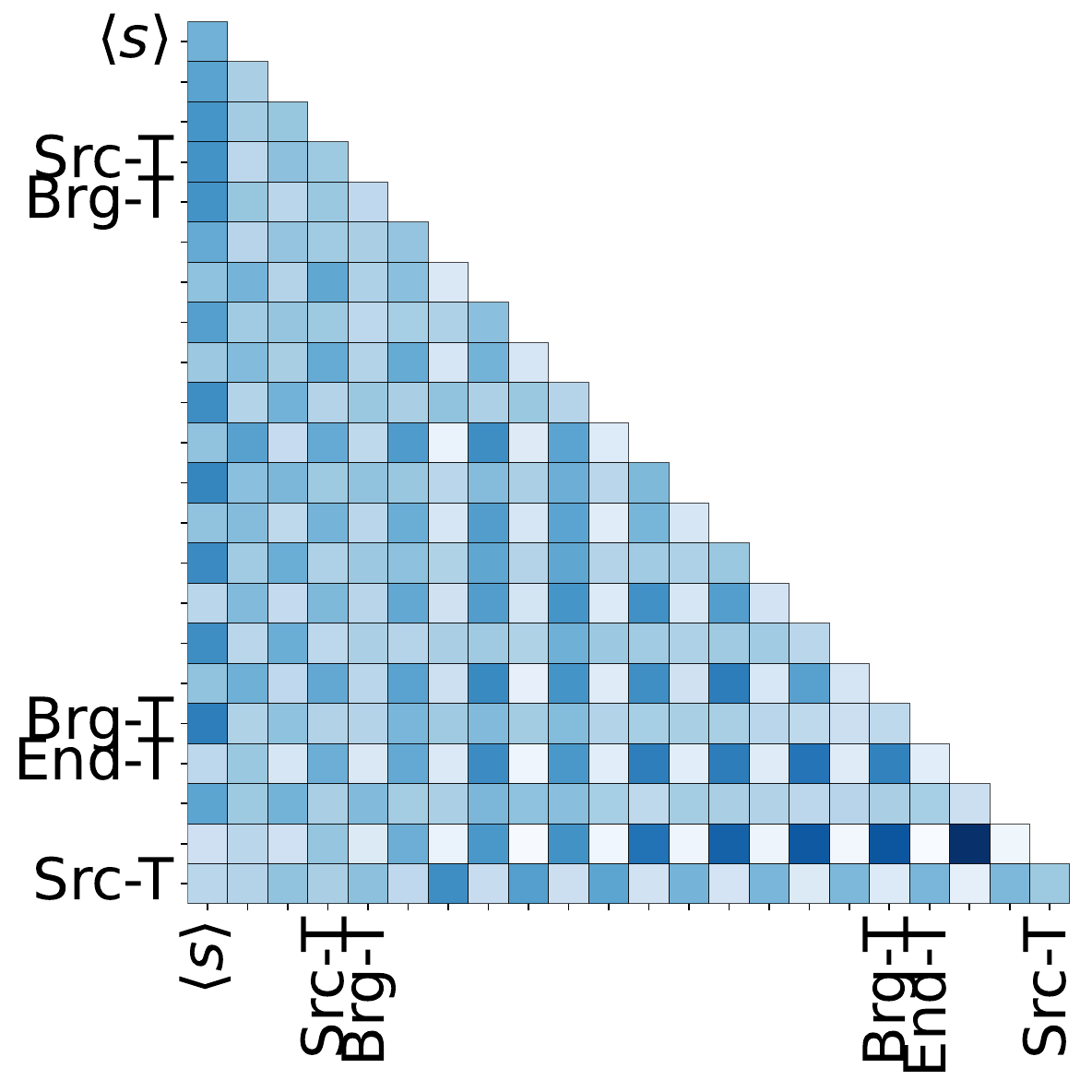}
    \end{subfigure}
    \hfill
    \begin{subfigure}{0.3\textwidth}
        \centering
        \caption{\small{Layer 3}}
        \label{fig:step800-layer-3}
        \includegraphics[width=\textwidth]{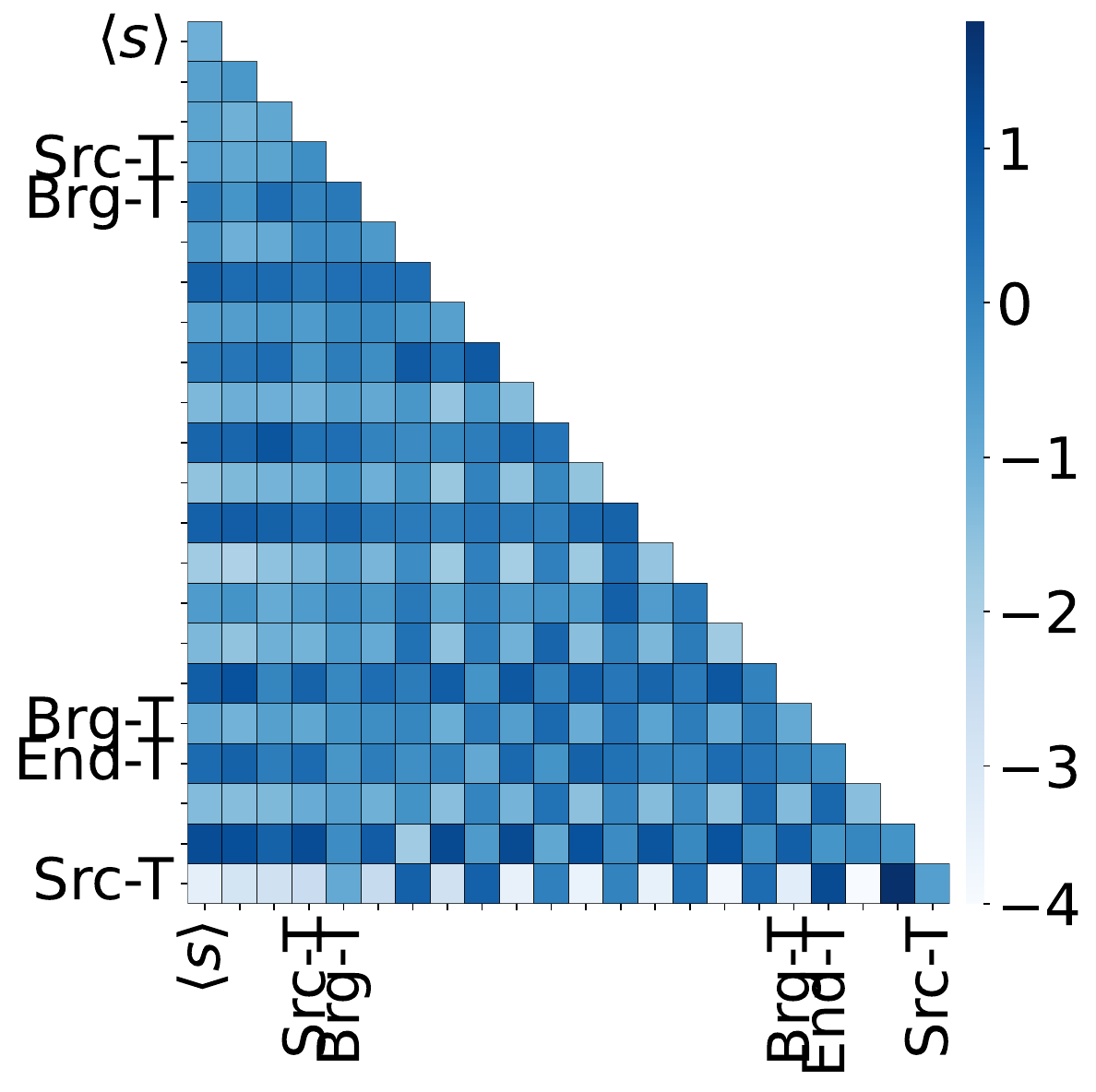}
    \end{subfigure}
    \caption{\small{\textbf{Attention logits heatmaps of the three-layer Transformer at the slow learning phase (training step 800)}:} \textit{Left (a):} The first layer. Chess-board-like pattern; \textit{Middle (b):} The second layer. The parent tokens have uniform attention on all preceding tokens. Each child token still \textit{uniformly} attends to \textit{all} parent tokens appearing in the preceding premises; \textit{Right (c):} The third layer. The query token uniformly attends to \textit{all} child tokens in the preceding premises. The query token retrieves the information from all \text{\Ed}~and \Brg~tokens. The logit lens could give a complete explanation for the random guessing, as shown in Figure~\ref{fig:random-guessing}.}
\end{figure}

\paragraph{The first layer.} The first attention layer shows chess-board-like patterns. Each parent token attend to \textit{all} child tokens appearing in the preceding premises, and vice versa, forming the alternating bright-dark grids on the heatmap, as shown in Figure~\ref{fig:step800-layer-1}. In particular, \text{\Ed}{} tokens (including \targetend{} and \nontargetend{} tokens) retrieve information from \Src{}~and \Brg{}~tokens.

\paragraph{The second layer.}
In the second layer, we observe a different information retrieval pattern, as presented in Figure~\ref{fig:step800-layer-2}. 
Here, each parent token changes to attend to all preceding tokens uniformly. Each child token still \textit{uniformly} attends to \textit{all} parent tokens appearing in the preceding premises. 
Crucially, for every child token, these attention logits are \textit{approximately equal} in magnitude. 
This uniform attention suggests that, at this intermediate stage, the model indiscriminately aggregates information from all prior parent tokens in the context without distinguishing between relevant and irrelevant tokens.

\paragraph{The third layer.}
The third layer attention logits in Figure~\ref{fig:step800-layer-3} show that the query token attends to all preceding child tokens, including all \Brg{} tokens in \Src{}-to-\Brg{} premises and all \text{\Ed}{} tokens in \Brg{}-to-\text{\Ed}{} premises.

\paragraph{Mechanistic Explanation of random guessing.}
Surprisingly, despite the final layer query token attending to all preceding child tokens, the resulting next-token probabilities strongly favor \text{\Ed}{} tokens, resulting in the random guessing over all \text{\Ed}{} tokens instead of all child tokens of the premises (as \Brg{} tokens are also child tokens in some premises).

\begin{minipage}{0.4\textwidth}
\begin{texample}[\shortstack{One Distractor}]\label{example.random.guess}
\small{Context: \Bos{} \textcolor{blue}{\nontargetsource}  \textcolor{blue}{\nontargetbridge}  \textcolor{red}{\targetsource}  \textcolor{red}{\targetbridge}  \textcolor{blue}{\nontargetbridge} \textcolor{blue}{\nontargetend} \textcolor{red}{\targetbridge} \textcolor{red}{\targetend} \ \targetsource ?\\[5pt]
\Bos{}: Begin-of-sequence token.\\
\textcolor{red}{Red}: Tokens from the target chain.\\
\textcolor{blue}{Blue}: The distractor chain.\\
\targetsource: The query token.}
\end{texample}
\end{minipage}
\hfill
\begin{minipage}{0.58\textwidth}
\includegraphics[width=\textwidth]{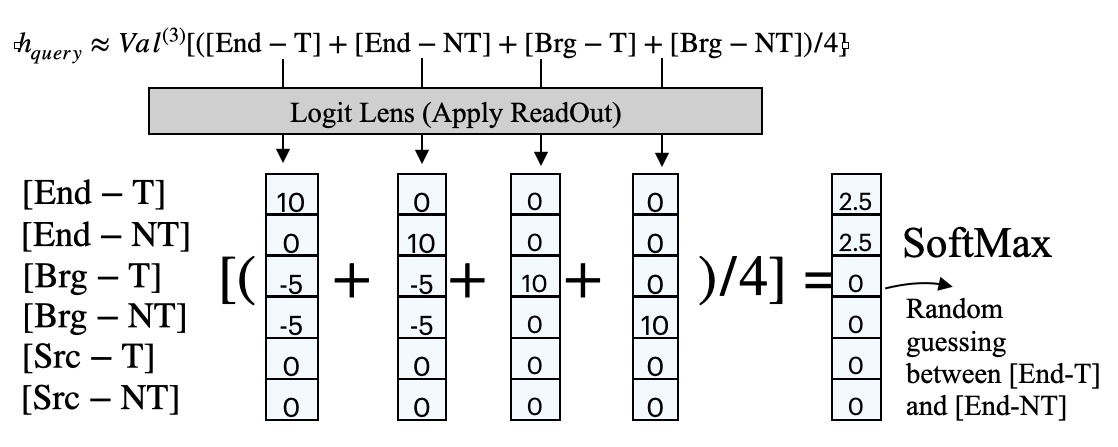}
\captionof{figure}{Illustration of the logit lens results of the query token at the layer 3 during the slow learning phase. The \text{\Ed}~tokens have positive entries at their own positions and negative entries with their preceding \Brg~tokens. These positive and negative values cancel each other out after summation.}
\label{fig:random-guessing}
\end{minipage}

The logit lens clearly illustrates this phenomenon. For visualization purposes, Figure~\ref{fig:random-guessing} presents the logit lens results with a simplified two-hop reasoning task (containing only one distractor) as shown in Example~\ref{example.random.guess}. The complete numerical results are provided in Section~\ref{sec:additional}. In Figure~\ref{fig:random-guessing}, each \text{\Ed}{} token has \textit{negative entries} associated with preceding parent tokens and \textit{positive entries} at its own position. 
This is probably associated with the fact that every child token attends to \textit{all preceding parent tokens} in the second layer.
Thus, the \Readout{} operation at the query token aggregates the value vectors equally.
Formally, this is
\begin{align}
    \textsc{ReadOut} (h_{\mathsf{query}})
    &\approx \frac{1}{|\textsc{child tokens}|} \sum_{i \in \textsc{Child tokens}} \textsc{ReadOut} (\textsc{Val}(i)) \notag \\
    &= \frac{1}{|\textsc{child tokens}|} \bigg[\sum_{i \in \textsc{Child tokens}} \mathbf{e}_i - 
    \sum_{j \in \textsc{Parent tokens}} a_j \mathbf{e}_j
    \bigg],
\end{align}
where $a_j > 0$ are some positive numbers, $\mathbf{e}_i$ is the vector with $i$-tn entry being one and others zero. $|\textsc{child tokens}|$ is the number of all preceding child tokens. 
Since some negative factors always appear at the \Src{} and \Brg{} tokens' entries, the most significant entries in $\textsc{ReadOut} (h_{\mathsf{query}})$ are all \text{\Ed}{} tokens.
After applying softmax, all other entries will become negligible, while similar positive magnitudes for all \text{\Ed}{} tokens yield approximately equal probabilities, resulting in random guesses over all \text{\Ed}{} tokens.

\subsection{Mechanistic interpretation for the structured learning phase}

In our experiments, after several hundred gradient steps, the three-layer Transformer successfully learns the two-hop tasks. To understand how this happens, we closely examine attention logits across each layer.

\begin{figure}[ht]
    \centering
    \begin{subfigure}{0.3\textwidth}
        \centering
        \caption{\small{Layer 1}}
        \label{fig:step10000-layer-1}
        \includegraphics[width=\textwidth]{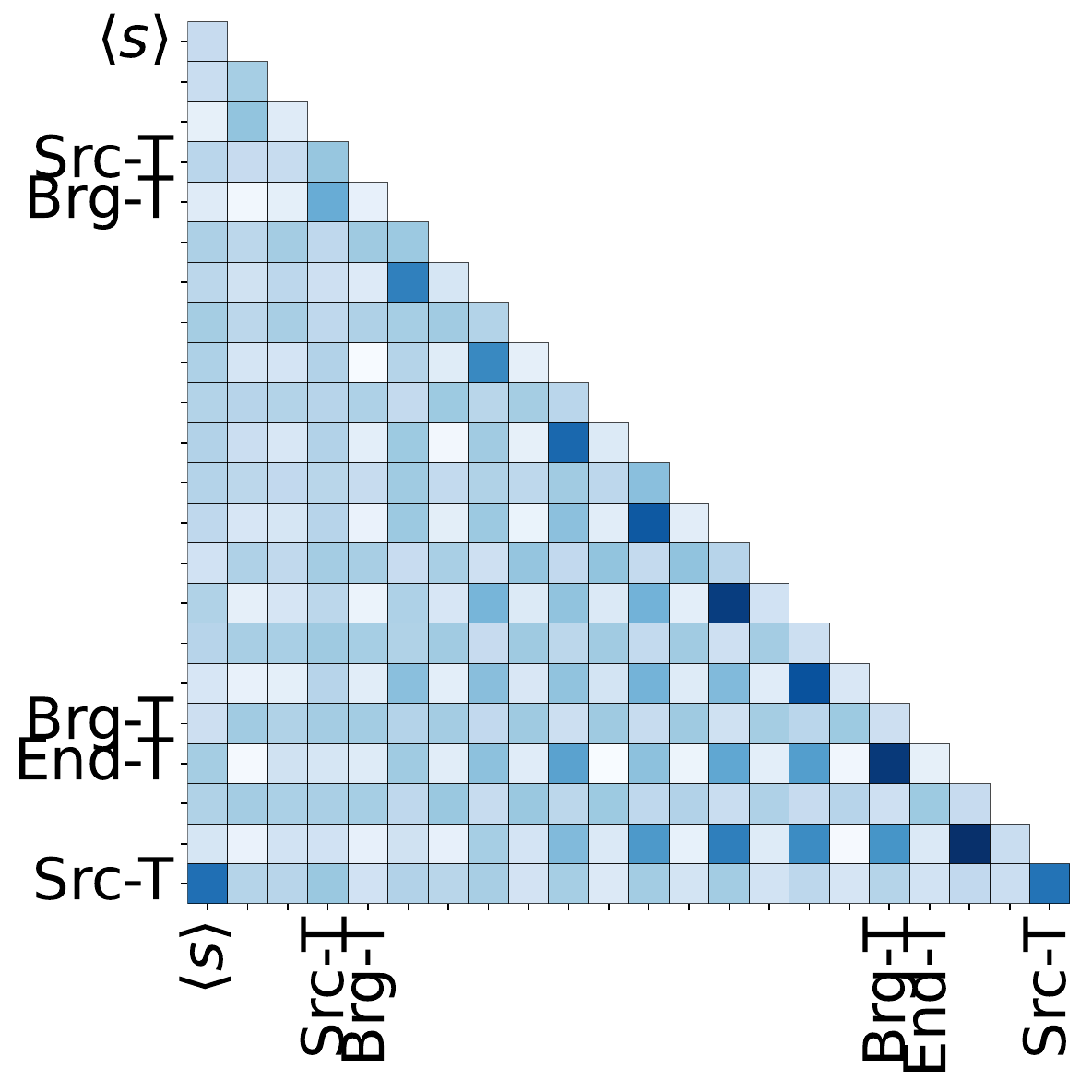}
    \end{subfigure}
    \hfill
    \begin{subfigure}{0.3\textwidth}
        \centering
        \caption{\small{Layer 2}}
        \label{fig:step10000-layer-2}
        \includegraphics[width=\textwidth]{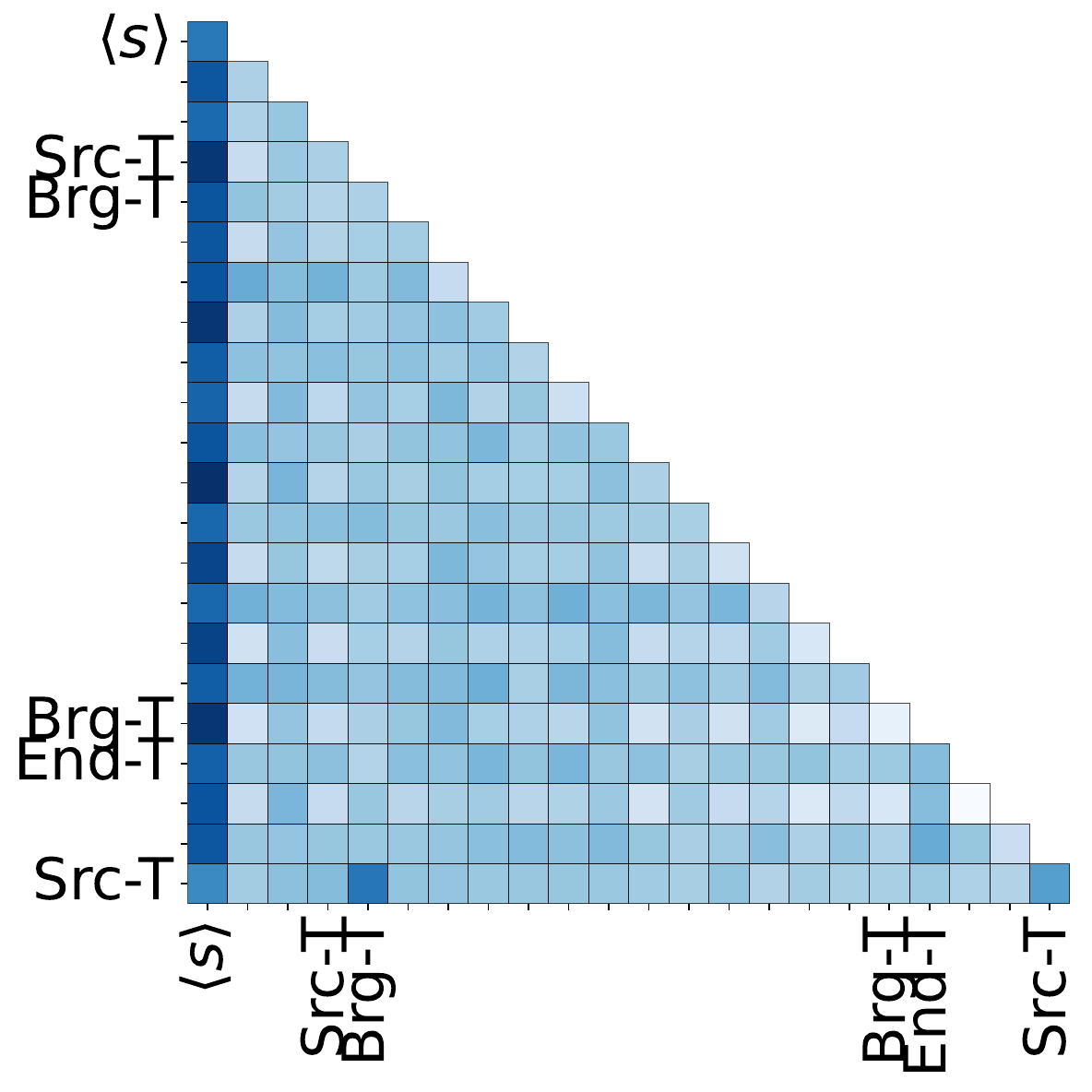}
    \end{subfigure}
    \hfill
    \begin{subfigure}{0.3\textwidth}
        \centering
        \caption{\small{Layer 3}}
        \label{fig:step10000-layer-3}
        \includegraphics[width=\textwidth]{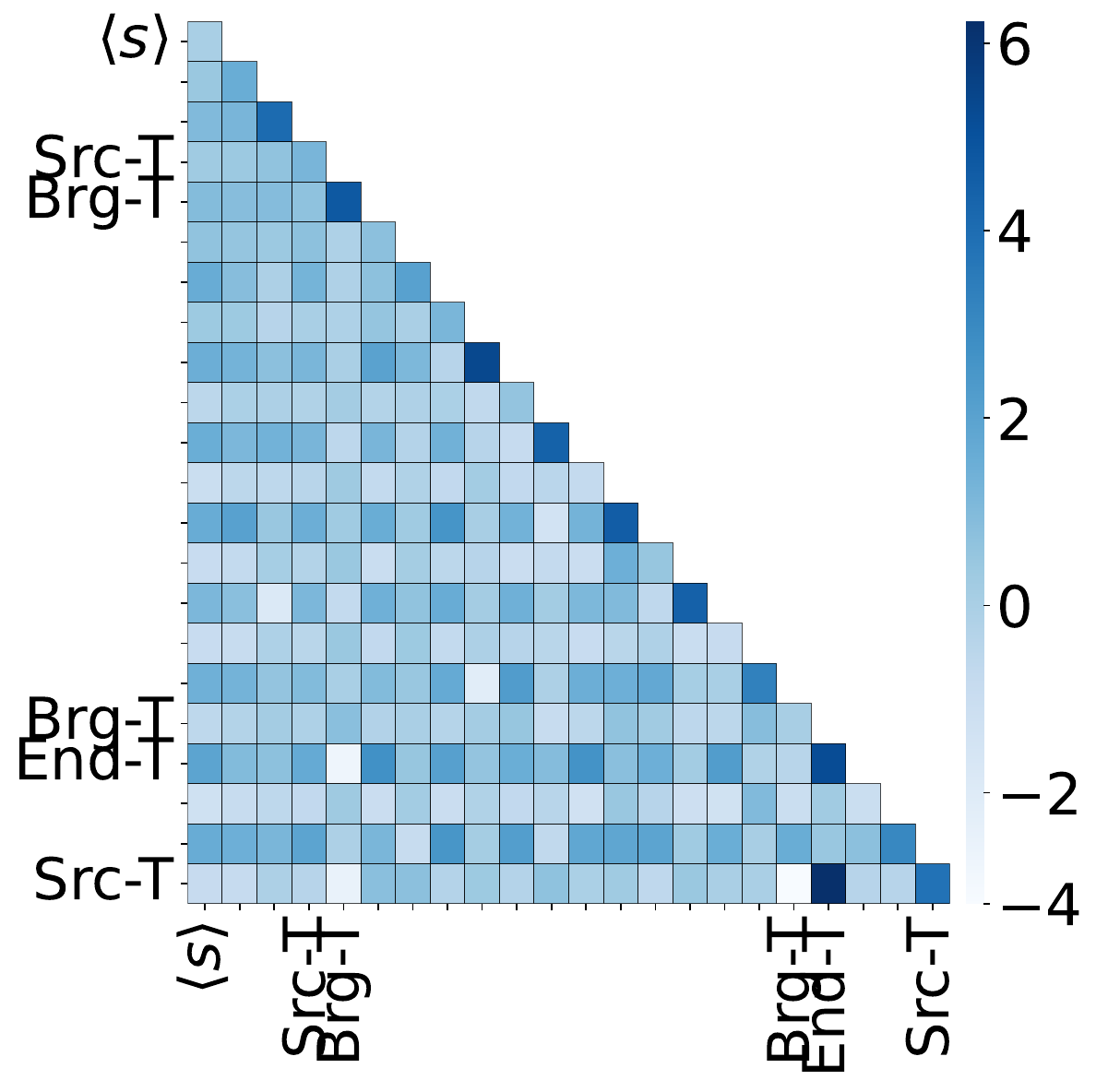}
    \end{subfigure}
    \caption{\small{\textbf{Attention logits heatmaps of the three-layer Transformer at the structured learning phase (training step 10000)}:} \textit{Left (a):} The first layer. Each child token strongly attends to its parent token; \textit{Middle (b):} The second layer. The query token strongly attends to the target bridge token (\targetbridge{}); \textit{Right (c):} The third layer. The query token strongly attends to the target end token (\targetend{}). The query retrieve the identity of the target end token (\targetend{}), enabling the correct next-token prediction.}
\end{figure}

\paragraph{The first layer.}
The first attention layer shows a clear token-copying mechanism. As shown in Figure~\ref{fig:step10000-layer-1}, the only significant entries in the attention map are those connecting the paired parent
token and the child token within every individual premise.
Each child token copies information directly from the paired parent token. For instance, the target bridge token (\targetbridge{}) retrieves information from the target source token (\targetsource{}), and the target end token (\targetend{}) similarly retrieves information from the target bridge token in \targetbridge{}-to-\targetend{} premise.

\paragraph{The second layer.}
In the second layer, the only significant attention occurs primarily between the query token and the target bridge token (\targetbridge{}). 
This strong attention indicates that the query token retrieves important information directly from the \targetbridge{} token. Since the \targetbridge{} token already contains information from its parent (the \targetsource{} token) due to the first-layer copying, the query token now possesses information from both \targetsource{} and \targetbridge{} tokens. 
This targeted retrieval is possible because the query and \targetbridge{} tokens share the same information (of the \targetsource{} token, since the \targetbridge{} token attends to \targetsource{} from the first layer, while the query token retrieves \targetsource{} in the second layer), allowing them to effectively attend to each other.


\paragraph{The Third Layer.}
In the final layer, attention exclusively connects the query token and the target end token (\targetend{}). 
This selective attention arises because both tokens share information from the target bridge token (\targetbridge{}): the \targetend{} token received this information directly from \targetbridge{} in the first layer, while the query token obtained it indirectly through the second layer. 
Consequently, the final attention mechanism precisely aligns the query token with the \targetend{} token, resulting in a next-token prediction that distinctly favors the \targetend{} token. 
This structured attention ensures the model’s predictions are both accurate and interpretable.

\paragraph{A simplified three-layer model.} To understand the relationship between the observed mechanisms and the loss training dynamics of the three-layer Transformer. We further build a three-parameter model that essentially captures both the observed mechanisms and the loss dynamics of Transformers, building strong connections between them. We relegate the details to Appendix~\ref{sec:three-param}.

\section{Conclusions}
\label{sec:conclusion}
In this paper, we study the underlying mechanism that transformer-based LLMs use to solve in-context two-hop reasoning tasks, especially in the presence of distracting information. We synthesize a novel dataset and find with Llama2-7B-Base that large pre-trained models may perform the uniform guessing mechanism on the two-hop reasoning task in context, and very few steps of fine-tuning suffice to teach the model to learn a correct mechanism. By carefully analyzing the training dynamics and fully reverse-engineering a three-layer Transformer, we identified the random guessing mechanism during the early training stages and the structured learning after a sharp phase transition. Our work could bring new insights into the internal reasoning mechanisms of LLMs. Extending our work to multi-hop reasoning and more LLMs would be important future work.

\newpage
\bibliographystyle{plainnat}
\bibliography{main}

\begin{thebibliography}{51}
\providecommand{\natexlab}[1]{#1}
\providecommand{\url}[1]{\texttt{#1}}
\expandafter\ifx\csname urlstyle\endcsname\relax
  \providecommand{\doi}[1]{doi: #1}\else
  \providecommand{\doi}{doi: \begingroup \urlstyle{rm}\Url}\fi

\bibitem[Achiam et~al.(2023)Achiam, Adler, Agarwal, Ahmad, Akkaya, Aleman, Almeida, Altenschmidt, Altman, Anadkat, et~al.]{achiam2023gpt}
Josh Achiam, Steven Adler, Sandhini Agarwal, Lama Ahmad, Ilge Akkaya, Florencia~Leoni Aleman, Diogo Almeida, Janko Altenschmidt, Sam Altman, Shyamal Anadkat, et~al.
\newblock Gpt-4 technical report.
\newblock \emph{arXiv preprint arXiv:2303.08774}, 2023.

\bibitem[Ahn et~al.(2024)Ahn, Cheng, Daneshmand, and Sra]{ahn2024transformers}
Kwangjun Ahn, Xiang Cheng, Hadi Daneshmand, and Suvrit Sra.
\newblock Transformers learn to implement preconditioned gradient descent for in-context learning.
\newblock \emph{Advances in Neural Information Processing Systems}, 36, 2024.

\bibitem[Aky{\"u}rek et~al.(2022)Aky{\"u}rek, Schuurmans, Andreas, Ma, and Zhou]{akyurek2022learning}
Ekin Aky{\"u}rek, Dale Schuurmans, Jacob Andreas, Tengyu Ma, and Denny Zhou.
\newblock What learning algorithm is in-context learning? investigations with linear models.
\newblock \emph{arXiv preprint arXiv:2211.15661}, 2022.

\bibitem[Allen-Zhu and Li(2023)]{allen2023physics}
Zeyuan Allen-Zhu and Yuanzhi Li.
\newblock Physics of language models: {P}art 1, context-free grammar.
\newblock \emph{arXiv preprint arXiv:2305.13673}, 2023.

\bibitem[Bai et~al.(2023)Bai, Chen, Wang, Xiong, and Mei]{bai2023transformers}
Yu~Bai, Fan Chen, Huan Wang, Caiming Xiong, and Song Mei.
\newblock Transformers as statisticians: Provable in-context learning with in-context algorithm selection.
\newblock \emph{Advances in neural information processing systems}, 36:\penalty0 57125--57211, 2023.

\bibitem[Belrose et~al.(2023)Belrose, Furman, Smith, Halawi, Ostrovsky, McKinney, Biderman, and Steinhardt]{belrose2023eliciting}
Nora Belrose, Zach Furman, Logan Smith, Danny Halawi, Igor Ostrovsky, Lev McKinney, Stella Biderman, and Jacob Steinhardt.
\newblock Eliciting latent predictions from transformers with the tuned lens.
\newblock \emph{arXiv preprint arXiv:2303.08112}, 2023.

\bibitem[Bietti et~al.(2023)Bietti, Cabannes, Bouchacourt, Jégou, and Bottou]{bietti2023birth}
Alberto Bietti, Vivien Cabannes, Diane Bouchacourt, Hervé Jégou, and Léon Bottou.
\newblock Birth of a transformer: A memory viewpoint, 2023.
\newblock URL \url{https://arxiv.org/abs/2306.00802}.

\bibitem[Bietti et~al.(2024)Bietti, Cabannes, Bouchacourt, Jegou, and Bottou]{bietti2024birth}
Alberto Bietti, Vivien Cabannes, Diane Bouchacourt, Herve Jegou, and Leon Bottou.
\newblock Birth of a transformer: A memory viewpoint.
\newblock \emph{Advances in Neural Information Processing Systems}, 36, 2024.

\bibitem[Biran et~al.(2024)Biran, Gottesman, Yang, Geva, and Globerson]{biran2024hopping}
Eden Biran, Daniela Gottesman, Sohee Yang, Mor Geva, and Amir Globerson.
\newblock Hopping too late: Exploring the limitations of large language models on multi-hop queries.
\newblock \emph{arXiv preprint arXiv:2406.12775}, 2024.

\bibitem[Brown et~al.(2020)Brown, Mann, Ryder, Subbiah, Kaplan, Dhariwal, Neelakantan, Shyam, Sastry, Askell, et~al.]{brown2020language}
Tom Brown, Benjamin Mann, Nick Ryder, Melanie Subbiah, Jared~D Kaplan, Prafulla Dhariwal, Arvind Neelakantan, Pranav Shyam, Girish Sastry, Amanda Askell, et~al.
\newblock Language models are few-shot learners.
\newblock \emph{Advances in neural information processing systems}, 33:\penalty0 1877--1901, 2020.

\bibitem[Charton(2022)]{charton2022my}
Fran{\c{c}}ois Charton.
\newblock What is my math transformer doing? {T}hree results on interpretability and generalization.
\newblock \emph{arXiv preprint arXiv:2211.00170}, 2022.

\bibitem[Chen et~al.(2024)Chen, Sheen, Wang, and Yang]{chen2024unveiling}
Siyu Chen, Heejune Sheen, Tianhao Wang, and Zhuoran Yang.
\newblock Unveiling induction heads: Provable training dynamics and feature learning in transformers.
\newblock \emph{arXiv preprint arXiv:2409.10559}, 2024.

\bibitem[Cobbe et~al.(2021)Cobbe, Kosaraju, Bavarian, Chen, Jun, Kaiser, Plappert, Tworek, Hilton, Nakano, et~al.]{cobbe2021training}
Karl Cobbe, Vineet Kosaraju, Mohammad Bavarian, Mark Chen, Heewoo Jun, Lukasz Kaiser, Matthias Plappert, Jerry Tworek, Jacob Hilton, Reiichiro Nakano, et~al.
\newblock Training verifiers to solve math word problems.
\newblock \emph{arXiv preprint arXiv:2110.14168}, 2021.

\bibitem[Conmy et~al.(2023)Conmy, Mavor-Parker, Lynch, Heimersheim, and Garriga-Alonso]{conmy2023towards}
Arthur Conmy, Augustine Mavor-Parker, Aengus Lynch, Stefan Heimersheim, and Adri{\`a} Garriga-Alonso.
\newblock Towards automated circuit discovery for mechanistic interpretability.
\newblock \emph{Advances in Neural Information Processing Systems}, 36:\penalty0 16318--16352, 2023.

\bibitem[Dubey et~al.(2024)Dubey, Jauhri, Pandey, Kadian, Al-Dahle, Letman, Mathur, Schelten, Yang, Fan, et~al.]{dubey2024llama}
Abhimanyu Dubey, Abhinav Jauhri, Abhinav Pandey, Abhishek Kadian, Ahmad Al-Dahle, Aiesha Letman, Akhil Mathur, Alan Schelten, Amy Yang, Angela Fan, et~al.
\newblock The llama 3 herd of models.
\newblock \emph{arXiv preprint arXiv:2407.21783}, 2024.

\bibitem[Elhage et~al.(2021)Elhage, Nanda, Olsson, Henighan, Joseph, Mann, Askell, Bai, Chen, Conerly, DasSarma, Drain, Ganguli, Hatfield-Dodds, Hernandez, Jones, Kernion, Lovitt, Ndousse, Amodei, Brown, Clark, Kaplan, McCandlish, and Olah]{elhage2021mathematical}
Nelson Elhage, Neel Nanda, Catherine Olsson, Tom Henighan, Nicholas Joseph, Ben Mann, Amanda Askell, Yuntao Bai, Anna Chen, Tom Conerly, Nova DasSarma, Dawn Drain, Deep Ganguli, Zac Hatfield-Dodds, Danny Hernandez, Andy Jones, Jackson Kernion, Liane Lovitt, Kamal Ndousse, Dario Amodei, Tom Brown, Jack Clark, Jared Kaplan, Sam McCandlish, and Chris Olah.
\newblock A mathematical framework for transformer circuits, 2021.
\newblock URL \url{https://transformer-circuits.pub/2021/framework/index.html}.

\bibitem[Feng and Steinhardt(2023)]{feng2023language}
Jiahai Feng and Jacob Steinhardt.
\newblock How do language models bind entities in context?
\newblock \emph{arXiv preprint arXiv:2310.17191}, 2023.

\bibitem[Feng et~al.(2024)Feng, Russell, and Steinhardt]{feng2024extractive}
Jiahai Feng, Stuart Russell, and Jacob Steinhardt.
\newblock Extractive structures learned in pretraining enable generalization on finetuned facts, 2024.
\newblock URL \url{https://arxiv.org/abs/2412.04614}.

\bibitem[Garg et~al.(2022)Garg, Tsipras, Liang, and Valiant]{garg2022can}
Shivam Garg, Dimitris Tsipras, Percy~S Liang, and Gregory Valiant.
\newblock What can transformers learn in-context? a case study of simple function classes.
\newblock \emph{Advances in Neural Information Processing Systems}, 35:\penalty0 30583--30598, 2022.

\bibitem[Geva et~al.(2023)Geva, Bastings, Filippova, and Globerson]{geva2023dissecting}
Mor Geva, Jasmijn Bastings, Katja Filippova, and Amir Globerson.
\newblock Dissecting recall of factual associations in auto-regressive language models.
\newblock \emph{arXiv preprint arXiv:2304.14767}, 2023.

\bibitem[Guo et~al.(2025)Guo, Yang, Zhang, Song, Zhang, Xu, Zhu, Ma, Wang, Bi, et~al.]{guo2025deepseek}
Daya Guo, Dejian Yang, Haowei Zhang, Junxiao Song, Ruoyu Zhang, Runxin Xu, Qihao Zhu, Shirong Ma, Peiyi Wang, Xiao Bi, et~al.
\newblock Deepseek-r1: Incentivizing reasoning capability in llms via reinforcement learning.
\newblock \emph{arXiv preprint arXiv:2501.12948}, 2025.

\bibitem[Guo et~al.(2023)Guo, Hu, Mei, Wang, Xiong, Savarese, and Bai]{guo2023transformers}
Tianyu Guo, Wei Hu, Song Mei, Huan Wang, Caiming Xiong, Silvio Savarese, and Yu~Bai.
\newblock How do transformers learn in-context beyond simple functions? a case study on learning with representations.
\newblock \emph{arXiv preprint arXiv:2310.10616}, 2023.

\bibitem[Guo et~al.(2024)Guo, Pai, Bai, Jiao, Jordan, and Mei]{guo2024active}
Tianyu Guo, Druv Pai, Yu~Bai, Jiantao Jiao, Michael~I Jordan, and Song Mei.
\newblock Active-dormant attention heads: Mechanistically demystifying extreme-token phenomena in llms.
\newblock \emph{arXiv preprint arXiv:2410.13835}, 2024.

\bibitem[Hase et~al.(2024)Hase, Bansal, Kim, and Ghandeharioun]{hase2024does}
Peter Hase, Mohit Bansal, Been Kim, and Asma Ghandeharioun.
\newblock Does localization inform editing? surprising differences in causality-based localization vs. knowledge editing in language models.
\newblock \emph{Advances in Neural Information Processing Systems}, 36, 2024.

\bibitem[Hu et~al.(2022)Hu, Shen, Wallis, Allen-Zhu, Li, Wang, Wang, Chen, et~al.]{hu2022lora}
Edward~J Hu, Yelong Shen, Phillip Wallis, Zeyuan Allen-Zhu, Yuanzhi Li, Shean Wang, Lu~Wang, Weizhu Chen, et~al.
\newblock Lora: Low-rank adaptation of large language models.
\newblock \emph{ICLR}, 1\penalty0 (2):\penalty0 3, 2022.

\bibitem[Huang et~al.(2023)Huang, Cheng, and Liang]{huang2023context}
Yu~Huang, Yuan Cheng, and Yingbin Liang.
\newblock In-context convergence of transformers.
\newblock \emph{arXiv preprint arXiv:2310.05249}, 2023.

\bibitem[Lin et~al.(2023)Lin, Bai, and Mei]{lin2023transformers}
Licong Lin, Yu~Bai, and Song Mei.
\newblock Transformers as decision makers: Provable in-context reinforcement learning via supervised pretraining.
\newblock \emph{arXiv preprint arXiv:2310.08566}, 2023.

\bibitem[Liu et~al.(2022)Liu, Kitouni, Nolte, Michaud, Tegmark, and Williams]{liu2022towards}
Ziming Liu, Ouail Kitouni, Niklas~S Nolte, Eric Michaud, Max Tegmark, and Mike Williams.
\newblock Towards understanding grokking: An effective theory of representation learning.
\newblock \emph{Advances in Neural Information Processing Systems}, 35:\penalty0 34651--34663, 2022.

\bibitem[Meng et~al.(2022)Meng, Bau, Andonian, and Belinkov]{meng2022locating}
Kevin Meng, David Bau, Alex Andonian, and Yonatan Belinkov.
\newblock Locating and editing factual associations in gpt.
\newblock \emph{Advances in Neural Information Processing Systems}, 35:\penalty0 17359--17372, 2022.

\bibitem[Nanda et~al.(2023)Nanda, Chan, Lieberum, Smith, and Steinhardt]{nanda2023progress}
Neel Nanda, Lawrence Chan, Tom Lieberum, Jess Smith, and Jacob Steinhardt.
\newblock Progress measures for grokking via mechanistic interpretability.
\newblock \emph{arXiv preprint arXiv:2301.05217}, 2023.

\bibitem[Nichani et~al.(2024)Nichani, Damian, and Lee]{nichani2024transformers}
Eshaan Nichani, Alex Damian, and Jason~D Lee.
\newblock How transformers learn causal structure with gradient descent.
\newblock \emph{arXiv preprint arXiv:2402.14735}, 2024.

\bibitem[Olsson et~al.(2022)Olsson, Elhage, Nanda, Joseph, DasSarma, Henighan, Mann, Askell, Bai, Chen, et~al.]{olsson2022context}
Catherine Olsson, Nelson Elhage, Neel Nanda, Nicholas Joseph, Nova DasSarma, Tom Henighan, Ben Mann, Amanda Askell, Yuntao Bai, Anna Chen, et~al.
\newblock In-context learning and induction heads.
\newblock \emph{arXiv preprint arXiv:2209.11895}, 2022.

\bibitem[Reddy(2023)]{reddy2023mechanistic}
Gautam Reddy.
\newblock The mechanistic basis of data dependence and abrupt learning in an in-context classification task.
\newblock In \emph{The Twelfth International Conference on Learning Representations}, 2023.

\bibitem[Sanford et~al.(2024{\natexlab{a}})Sanford, Hsu, and Telgarsky]{sanford2024one}
Clayton Sanford, Daniel Hsu, and Matus Telgarsky.
\newblock One-layer transformers fail to solve the induction heads task.
\newblock \emph{arXiv preprint arXiv:2408.14332}, 2024{\natexlab{a}}.

\bibitem[Sanford et~al.(2024{\natexlab{b}})Sanford, Hsu, and Telgarsky]{sanford2024transformers}
Clayton Sanford, Daniel Hsu, and Matus Telgarsky.
\newblock Transformers, parallel computation, and logarithmic depth, 2024{\natexlab{b}}.
\newblock URL \url{https://arxiv.org/abs/2402.09268}.

\bibitem[Shao et~al.(2024)Shao, Wang, Zhu, Xu, Song, Bi, Zhang, Zhang, Li, Wu, et~al.]{shao2024deepseekmath}
Zhihong Shao, Peiyi Wang, Qihao Zhu, Runxin Xu, Junxiao Song, Xiao Bi, Haowei Zhang, Mingchuan Zhang, YK~Li, Y~Wu, et~al.
\newblock Deepseekmath: Pushing the limits of mathematical reasoning in open language models.
\newblock \emph{arXiv preprint arXiv:2402.03300}, 2024.

\bibitem[Shi et~al.(2024)Shi, Beltran-Velez, Nazaret, Zheng, Garriga-Alonso, Jesson, Makar, and Blei]{shi2024hypothesis}
Claudia Shi, Nicolas Beltran-Velez, Achille Nazaret, Carolina Zheng, Adri{\`a} Garriga-Alonso, Andrew Jesson, Maggie Makar, and David~M Blei.
\newblock Hypothesis testing the circuit hypothesis in llms.
\newblock \emph{arXiv preprint arXiv:2410.13032}, 2024.

\bibitem[Todd et~al.(2023)Todd, Li, Sharma, Mueller, Wallace, and Bau]{todd2023function}
Eric Todd, Millicent~L Li, Arnab~Sen Sharma, Aaron Mueller, Byron~C Wallace, and David Bau.
\newblock Function vectors in large language models.
\newblock \emph{arXiv preprint arXiv:2310.15213}, 2023.

\bibitem[Von~Oswald et~al.(2023)Von~Oswald, Niklasson, Randazzo, Sacramento, Mordvintsev, Zhmoginov, and Vladymyrov]{von2023transformers}
Johannes Von~Oswald, Eyvind Niklasson, Ettore Randazzo, Jo{\~a}o Sacramento, Alexander Mordvintsev, Andrey Zhmoginov, and Max Vladymyrov.
\newblock Transformers learn in-context by gradient descent.
\newblock In \emph{International Conference on Machine Learning}, pages 35151--35174. PMLR, 2023.

\bibitem[Wang et~al.(2024{\natexlab{a}})Wang, Yue, Su, and Sun]{wang2024grokked}
Boshi Wang, Xiang Yue, Yu~Su, and Huan Sun.
\newblock Grokked transformers are implicit reasoners: A mechanistic journey to the edge of generalization.
\newblock \emph{arXiv preprint arXiv:2405.15071}, 2024{\natexlab{a}}.

\bibitem[Wang et~al.(2022)Wang, Variengien, Conmy, Shlegeris, and Steinhardt]{wang2022interpretability}
Kevin Wang, Alexandre Variengien, Arthur Conmy, Buck Shlegeris, and Jacob Steinhardt.
\newblock Interpretability in the wild: a circuit for indirect object identification in {GPT}-2 small.
\newblock \emph{arXiv preprint arXiv:2211.00593}, 2022.

\bibitem[Wang et~al.(2024{\natexlab{b}})Wang, Yu, E, and Wu]{wang2024transformers}
Mingze Wang, Ruoxi Yu, Weinan E, and Lei Wu.
\newblock How transformers implement induction heads: Approximation and optimization analysis, 2024{\natexlab{b}}.
\newblock URL \url{https://arxiv.org/abs/2410.11474}.

\bibitem[Wei et~al.(2022)Wei, Wang, Schuurmans, Bosma, Xia, Chi, Le, Zhou, et~al.]{wei2022chain}
Jason Wei, Xuezhi Wang, Dale Schuurmans, Maarten Bosma, Fei Xia, Ed~Chi, Quoc~V Le, Denny Zhou, et~al.
\newblock Chain-of-thought prompting elicits reasoning in large language models.
\newblock \emph{Advances in neural information processing systems}, 35:\penalty0 24824--24837, 2022.

\bibitem[Xie et~al.(2021)Xie, Raghunathan, Liang, and Ma]{xie2021explanation}
Sang~Michael Xie, Aditi Raghunathan, Percy Liang, and Tengyu Ma.
\newblock An explanation of in-context learning as implicit bayesian inference.
\newblock \emph{arXiv preprint arXiv:2111.02080}, 2021.

\bibitem[Yang et~al.(2024{\natexlab{a}})Yang, Yang, Zhang, Hui, Zheng, Yu, Li, Liu, Huang, Wei, et~al.]{yang2024qwen2}
An~Yang, Baosong Yang, Beichen Zhang, Binyuan Hui, Bo~Zheng, Bowen Yu, Chengyuan Li, Dayiheng Liu, Fei Huang, Haoran Wei, et~al.
\newblock Qwen2. 5 technical report.
\newblock \emph{arXiv preprint arXiv:2412.15115}, 2024{\natexlab{a}}.

\bibitem[Yang et~al.(2024{\natexlab{b}})Yang, Gribovskaya, Kassner, Geva, and Riedel]{yang2024latent}
Sohee Yang, Elena Gribovskaya, Nora Kassner, Mor Geva, and Sebastian Riedel.
\newblock Do large language models latently perform multi-hop reasoning?, 2024{\natexlab{b}}.
\newblock URL \url{https://arxiv.org/abs/2402.16837}.

\bibitem[Zhang et~al.(2024)Zhang, Frei, and Bartlett]{zhang2024trained}
Ruiqi Zhang, Spencer Frei, and Peter~L Bartlett.
\newblock Trained transformers learn linear models in-context.
\newblock \emph{Journal of Machine Learning Research}, 25\penalty0 (49):\penalty0 1--55, 2024.

\bibitem[Zhang et~al.(2022)Zhang, Backurs, Bubeck, Eldan, Gunasekar, and Wagner]{zhang2022unveiling}
Yi~Zhang, Arturs Backurs, S{\'e}bastien Bubeck, Ronen Eldan, Suriya Gunasekar, and Tal Wagner.
\newblock Unveiling transformers with {LEGO}: A synthetic reasoning task.
\newblock \emph{arXiv preprint arXiv:2206.04301}, 2022.

\bibitem[Zhong et~al.(2023)Zhong, Wu, Manning, Potts, and Chen]{zhong2023mquake}
Zexuan Zhong, Zhengxuan Wu, Christopher~D Manning, Christopher Potts, and Danqi Chen.
\newblock Mquake: Assessing knowledge editing in language models via multi-hop questions.
\newblock \emph{arXiv preprint arXiv:2305.14795}, 2023.

\bibitem[Zhu et~al.(2024)Zhu, Huang, Zhang, Jordan, Jiao, Tian, and Russell]{zhu2024towards}
Hanlin Zhu, Baihe Huang, Shaolun Zhang, Michael Jordan, Jiantao Jiao, Yuandong Tian, and Stuart~J Russell.
\newblock Towards a theoretical understanding of the'reversal curse'via training dynamics.
\newblock \emph{Advances in Neural Information Processing Systems}, 37:\penalty0 90473--90513, 2024.

\bibitem[Zhu and Li(2023)]{zhu2023physics}
Zeyuan~Allen Zhu and Yuanzhi Li.
\newblock Physics of language models: {P}art 3.1, knowledge storage and extraction.
\newblock \emph{arXiv preprint arXiv:2309.14316}, 2023.

\end{thebibliography}

\newpage
\appendix
\section{Additional Details in LLM Experiments}
\label{sec:llm}
\subsection{LLM Evaluation Details}
We design six two-hop reasoning templates with the aid of Deepseek-R1 \citep{guo2025deepseek}. We choose common male and female names. To avoid interference by prior knowledge, we generate fake locations and biological categories.

\begin{tcolorbox}[title=Templates for Two-hop Reasoning with Distractors]\label{template.two.hop.with.distractor}
\textbf{Family relations}: "[A] is the mother of [B]. [B] is the mother of [C]. Therefore, [A] is the grandmother of"\\
"[A] is the father of [B]. [B] is the father of [C]. Therefore, [A] is the grandfather of",

\textbf{Geographical relations}: "[A] is a city in the state of [B]. The state of [B] is part of the country [C]. Therefore, [A] is located in"\\
"[A] lives in [B]. People in [B] speak [C]. Therefore, [A] speaks",

\textbf{Biological relations}: "[A] is a species in the genus [B]. The genus [B] belongs to the family [C]. Therefore, [A] is classified under the family",

\textbf{Arithmetic relations}: "[A] follows the time zone of [B]. [B] is three hours ahead of [C]. Therefore, [A] is three hours ahead of",
\end{tcolorbox}

\begin{tcolorbox}[title=Entity names]
\textbf{Locations}: "Zorvath", "Tyseria", "Kryo", "Vynora", "Quellion", "Dras", "Luminax", "Vesperon", "Noctari", "Xyphodon", "Glacidae", "Ophirion", "Eryndor", "Solmyra", "Umbrithis", "Balthorien", "Ytheris", "Fendrel", "Havroth", "Marendor"

\textbf{Biological taxonomy}: "Fluxilus", "Varnex", "Dranthidae", "Zynthor", "Gryvus", "Myralin", "Thalorium", "Zephyra", "Aerinth", "Xyphodon", "Kryostis", "Glacidae", "Borithis", "Chrysalix", "Noctilura", "Phorvian", "Seraphid", "Uthrelin", "Eldrinth", "Yvorith"

\textbf{Languages}: "English", "Spanish", "Mandarin", "Hindi", "Arabic", "French", "German", "Japanese", "Portuguese", "Russian", "Korean", "Italian", "Turkish", "Dutch", "Swedish", "Polish", "Hebrew", "Greek", "Bengali", "Thai"

\textbf{Names}: "Ben", "Jack", "Luke", "Mark", "Paul", "John", "Tom", "Sam", "Joe", "Max", "Amy", "Emma", "Anna", "Grace",  "Kate", "Lucy", "Sarah", "Alice", "Alex", "Ruby"
\end{tcolorbox}

\subsection{LLM Finetuning Details}
We train four models on two-hop reasoning with
\textbf{one} distractor using LoRA \cite{hu2022lora}. We list the training configuration details in Table~\ref{tab:llm-ft-details}. The  experiments were completed within one hour utilizing four NVIDIA A100 GPUs.

\begin{table}[htbp]
\centering
\caption{LLM Fine-tuning with LoRA Configuration}
\label{tab:llm-ft-details}
\begin{tabular}{>{\raggedright\arraybackslash}p{4cm} >{\raggedright\arraybackslash}p{4cm}}
\toprule
\rowcolor{headercolor} \textbf{Setting} & \textbf{Value} \\
\midrule
\rowcolor{sectioncolor} \multicolumn{2}{l}{\textbf{Model Configuration}} \\
Model Names & Qwen2.5-7B, OLMo-7B, Llama2-7B, Llama3.1-8B \\
Precision & fp16 \\
\midrule
\rowcolor{sectioncolor} \multicolumn{2}{l}{\textbf{LoRA Configuration}} \\
LoRA Dimension (r) & 16 \\
LoRA Alpha Parameter & 32 \\
LoRA Dropout & 0.05 \\
Bias & None \\
Target Modules & ["q\_proj", "k\_proj", "v\_proj", "o\_proj"] \\
\midrule

\rowcolor{sectioncolor} \multicolumn{2}{l}{\textbf{Data Processing}} \\
Distractor Numbers & 2 \\
Maximum Sequence Length & 128 \\
\midrule

\rowcolor{sectioncolor} \multicolumn{2}{l}{\textbf{Optimization}} \\
Optimizer & AdamW \\
Learning rate & $2*10^{-4}$ \\
Learning rate scheduler & cosine \\
Warmup ratio & 0.03 \\
Weight decay & 0.01 \\
Epochs & 3 \\
Batch size & 16 \\
\bottomrule
\end{tabular}
\end{table}

\subsection{Evaluation results for more models}
We evaluate more LLMs on the two-hop reasoning with distractors before or after the finetuning. Tables~\ref{tab:qwen-random-guess}, \ref{tab:llama3.1-random-guess}, and~\ref{tab:olmo-random-guess} present the average probabilities on the answer and distractors.

\begin{table}[h]
\begin{center}
\caption{\small{
Performance comparison of Qwen2.5-7b base and fine-tuned models on two-hop reasoning tasks with varying numbers of distractors. 
$K=1$ indicates the scenario without distractors. 
The \targetend{} and \nontargetend{} rows report the average probabilities assigned to the first token of the target end entity and non-target end entities, respectively. 
For cases with multiple distractors, the reported probabilities for non-target entities are averaged first over all \nontargetend{} within each context, then across all contexts. 
The values in the parentheses are the standard errors.
} 
}
\vspace{1mm}
\resizebox{1\textwidth}{!}{
\begin{tabular}{c|c|ccccc}
\toprule[1pt]
\midrule
\multicolumn{1}{c|}{\multirow{2}{*}{\textbf{Models}}} &
\multicolumn{1}{c|}{\multirow{2}{*}{\textbf{Next Tokens}}} & 
\multicolumn{5}{c}{\textbf{The Number of Reasoning Chains in the Context $(K)$}} \\
\cmidrule{3-7}
& & \begin{tabular}{c} 1 \end{tabular}
& \begin{tabular}{c} 2 \end{tabular}
& \begin{tabular}{c} 3 \end{tabular}
& \begin{tabular}{c} 4 \end{tabular}
& \begin{tabular}{c} 5 \end{tabular}\\ 
\midrule
\multicolumn{1}{c|}{\multirow{2}{*}{\textbf{Qwen2.5-7B}}} & \targetend & 0.80 (0.01) & 0.29 (0.01) & 0.26 (0.01) & 0.26 (0.01) & 0.24 (0.01) \\
 & \nontargetend & NA (NA) & 0.27 (0.01) & 0.10 (0.00) & 0.06 (0.00) & 0.05 (0.00) \\
\midrule
\multicolumn{1}{c|}{\multirow{2}{*}{\shortstack{\textbf{Fine-tuned}\\ \textbf{on $K=2$}}}} & \targetend & 1.00 (0.00) & 1.00 (0.00) & 1.00 (0.00) & 1.00 (0.00) & 0.99 (0.00) \\
 & \nontargetend & NA (NA) & 0.00 (0.00) & 0.01 (0.00) & 0.00 (0.00) & 0.00 (0.00) \\
\midrule
\bottomrule
\end{tabular}
}
\label{tab:qwen-random-guess}
\end{center}
\end{table}

\begin{table}[h]
\begin{center}
\caption{\small{
Performance comparison of Llama3.1-8B base and fine-tuned models on two-hop reasoning tasks with varying numbers of distractors. 
$K=1$ indicates the scenario without distractors. 
The \targetend{} and \nontargetend{} rows report the average probabilities assigned to the first token of the target end entity and non-target end entities, respectively. 
For cases with multiple distractors, the reported probabilities for non-target entities are averaged first over all \nontargetend{} within each context, then across all contexts. 
The values in the parentheses are the standard errors.
} 
}
\vspace{1mm}
\resizebox{1\textwidth}{!}{
\begin{tabular}{c|c|ccccc}
\toprule[1pt]
\midrule
\multicolumn{1}{c|}{\multirow{2}{*}{\textbf{Models}}} &
\multicolumn{1}{c|}{\multirow{2}{*}{\textbf{Next Tokens}}} & 
\multicolumn{5}{c}{\textbf{The Number of Reasoning Chains in the Context $(K)$}} \\
\cmidrule{3-7}
& & \begin{tabular}{c} 1 \end{tabular}
& \begin{tabular}{c} 2 \end{tabular}
& \begin{tabular}{c} 3 \end{tabular}
& \begin{tabular}{c} 4 \end{tabular}
& \begin{tabular}{c} 5 \end{tabular}\\ 
\midrule
\multicolumn{1}{c|}{\multirow{2}{*}{\textbf{Llama3-8b}}} & \targetend & 0.90 (0.00) & 0.40 (0.01) & 0.32 (0.01) & 0.29 (0.01) & 0.27 (0.01) \\
 & \nontargetend & NA (NA) & 0.37 (0.01) & 0.19 (0.00) & 0.13 (0.00) & 0.10 (0.00) \\
\midrule
\multicolumn{1}{c|}{\multirow{2}{*}{\shortstack{\textbf{Fine-tuned}\\ \textbf{on $K=2$}}}} & \targetend & 1.00 (0.00) & 1.00 (0.00) & 1.00 (0.00) & 1.00 (0.00) & 1.00 (0.00) \\
 & \nontargetend & NA (NA) & 0.00 (0.00) & 0.00 (0.00) & 0.00 (0.00) & 0.00 (0.00) \\
\midrule
\bottomrule
\end{tabular}
}
\label{tab:llama3.1-random-guess}
\end{center}
\end{table}

\begin{table}[h]
\begin{center}
\caption{\small{
Performance comparison of OLMo-7B base and fine-tuned models on two-hop reasoning tasks with varying numbers of distractors. 
$K=1$ indicates the scenario without distractors. 
The \targetend{} and \nontargetend{} rows report the average probabilities assigned to the first token of the target end entity and non-target end entities, respectively. 
For cases with multiple distractors, the reported probabilities for non-target entities are averaged first over all \nontargetend{} within each context, then across all contexts. 
The values in the parentheses are the standard errors.
} 
}
\vspace{1mm}
\resizebox{1\textwidth}{!}{
\begin{tabular}{c|c|ccccc}
\toprule[1pt]
\midrule
\multicolumn{1}{c|}{\multirow{2}{*}{\textbf{Models}}} &
\multicolumn{1}{c|}{\multirow{2}{*}{\textbf{Next Tokens}}} & 
\multicolumn{5}{c}{\textbf{The Number of Reasoning Chains in the Context $(K)$}} \\
\cmidrule{3-7}
& & \begin{tabular}{c} 1 \end{tabular}
& \begin{tabular}{c} 2 \end{tabular}
& \begin{tabular}{c} 3 \end{tabular}
& \begin{tabular}{c} 4 \end{tabular}
& \begin{tabular}{c} 5 \end{tabular}\\ 
\midrule
\multicolumn{1}{c|}{\multirow{2}{*}{\textbf{OLMo}}} & \targetend & 0.72 (0.01) & 0.37 (0.01) & 0.25 (0.01) & 0.18 (0.01) & 0.14 (0.00) \\
 & \nontargetend & NA (NA) & 0.32 (0.01) & 0.19 (0.01) & 0.14 (0.00) & 0.11 (0.00) \\
\midrule
\multicolumn{1}{c|}{\multirow{2}{*}{\shortstack{\textbf{Fine-tuned}\\ \textbf{on $K=2$}}}} & \targetend & 1.00 (0.00) & 1.00 (0.00) & 0.66 (0.01) & 0.57 (0.01) & 0.50 (0.01) \\
 & \nontargetend & NA (NA) & 0.00 (0.00) & 0.17 (0.01) & 0.14 (0.01) & 0.12 (0.01) \\
\midrule
\bottomrule
\end{tabular}
}
\label{tab:olmo-random-guess}
\end{center}
\end{table}

\section{Additional Details in Three-layer Transformer Experiments}
\label{sec:additional}
\subsection{Additional training details}
We list the training details in Table~\ref{tab:tf-train-params}.

\begin{table}[htbp]
\centering
\caption{Transformer Training Configuration}
\label{tab:tf-train-params}
\begin{tabular}{>{\raggedright\arraybackslash}p{5cm} >{\raggedright\arraybackslash}p{6cm}}
\toprule
\rowcolor{headercolor} \textbf{Setting} & \textbf{Value/Default} \\
\midrule
\rowcolor{sectioncolor} \multicolumn{2}{l}{\textbf{Model Architecture}} \\
Embedding Type & Positional embeddings \\
Normalization & Pre-layer normalization \\
MLP Activation & ReLU in \(\text{mlp}\) \\
\midrule

\rowcolor{sectioncolor} \multicolumn{2}{l}{\textbf{Optimization Parameters}} \\
Optimizer & Adam \\
Learning Rate & 0.0003 (fixed) \\
$\beta_1$ & 0.9 \\
$\beta_2$ & 0.99 \\
$\epsilon$ & $10^{-8}$ \\
Weight Decay & 0.01 \\
\midrule

\rowcolor{sectioncolor} \multicolumn{2}{l}{\textbf{SGD Configuration}} \\
Learning Rate (3-parameter model) & 0.1 \\
\midrule

\rowcolor{sectioncolor} \multicolumn{2}{l}{\textbf{Training Details}} \\
Batch Size ($B$) & 512 \\
Sequence Length ($N$) & 23 \\
Training Steps & 10,000 \\
Seeds & Results are consistent across different random seeds \\
\bottomrule
\end{tabular}
\end{table}

\subsection{Logit Lens Results}
\begin{figure}
    \centering
\includegraphics[width=0.5\linewidth]{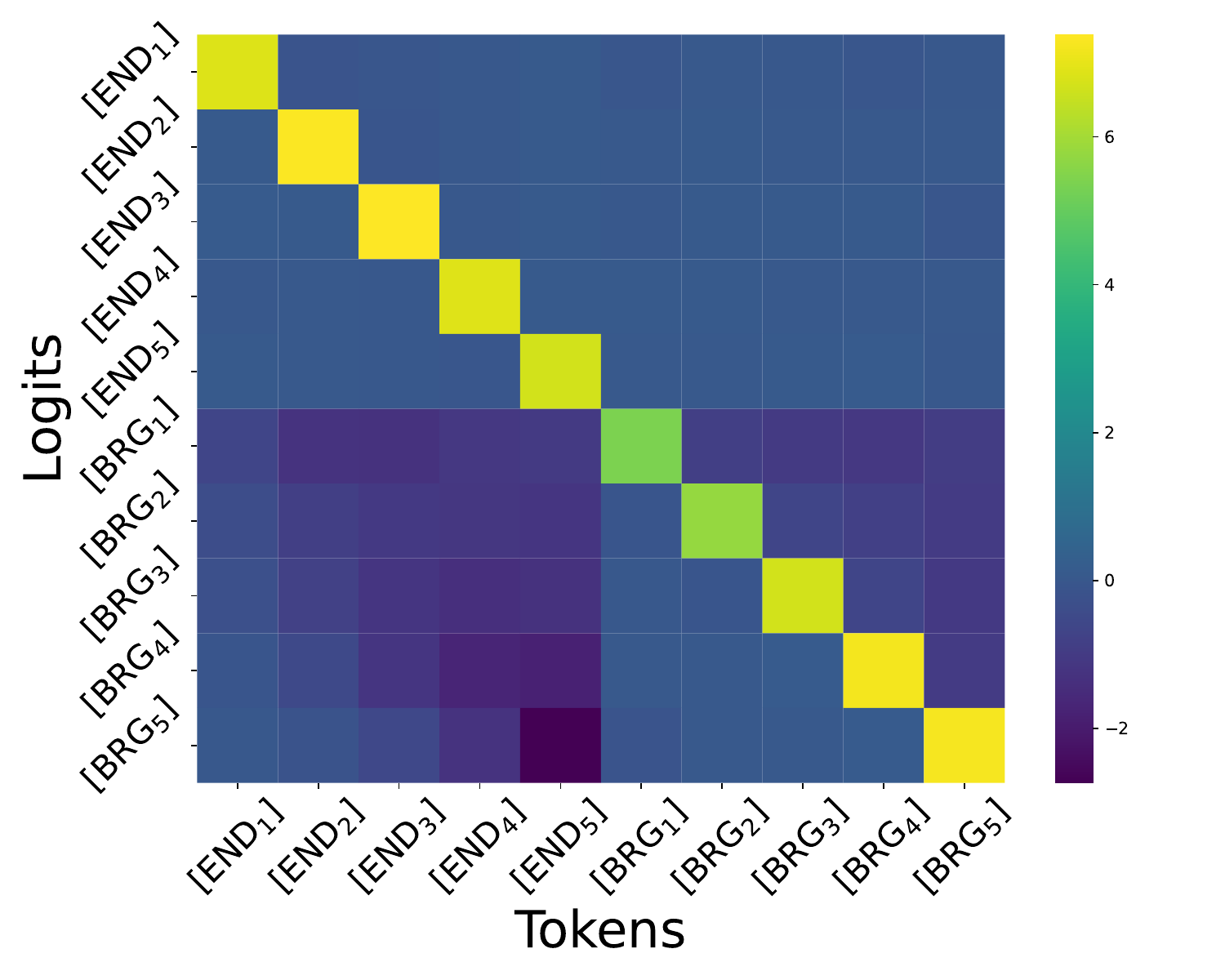}
    \caption{\textbf{Logit lens of the value states at the layer $3$}: We use the model trained on step $800$. The results are averaged over $256$ sequences. The $y$-axis represents the entries on the logit lens output. The $x$-axis represents tokens. The bright color indicates larger value, and blue color indicates negative values. The bright diagonal line shows that all tokens have value states that strongly support predicting themselves. The left bottom blue part indicates that $\text{\Ed}$ tokens have negative values for $\Brg$ tokens.}
    \label{fig:logit-lens-layer-3}
\end{figure}

We present the details for logit lens results. Suppose that given any input sequence, the value state of token $t$ at layer $3$ is $\textsc{value}^{(3)}$. Suppose the projection matrix in the attention layer $3$ is $\mathbf{O}^{(3)}$, the layer norm before the \Readout~matrix is $\textsc{LayerNorm}$. For any token $t\in \{$ \Brg[1], $\ldots$, \Brg[5], \text{\Ed}[1], $\ldots$, \text{\Ed}[5] $\}$, using the logit lens, we compute
\begin{equation*}
\textsc{Output-Logits}(t) = \Readout (\textsc{LayerNorm} (\mathbf{O}^{(3)} \textsc{value}^{(3)}(t))).
\end{equation*}
We focus on the entries on the $\textsc{Output-Logits}(t)$ corresponding to tokens in $t\in \{$ \Brg[1], $\ldots$, \Brg[5], \text{\Ed}[1], $\ldots$, \text{\Ed}[5] $\}$. Figure~\ref{fig:logit-lens-layer-3} shows their logit lens results computed at the model trained with 800 steps. All tokens have positive logits for themselves. $\text{\Ed}$ tokens have negative logits on $\Brg$ tokens. $\Brg$ tokens have approximately zero logits on $\text{\Ed}$ tokens.

\section{The Three-Parameter Model}
\label{sec:three-param}

\paragraph{``Causal'' hypotheses based on observations.} In Section~\ref{sec:tf-twohop}, we observe two stages along the training dynamics of the three-layer transformer. In this section, we aim to build a ``causal'' relationship between the observed mechanisms and the training dynamics. The two ``causal'' hypotheses are:
\begin{hypothesis}\label{hyp:random-guess}
The formation of the \textit{random guessing mechanism} causes the slow learning phase (0-800 steps).
\end{hypothesis}
\begin{hypothesis}\label{hyp:sequential-query}
The formation of the \textit{sequential query mechanism} causes the abrupt phase transition (800-10000 steps).
\end{hypothesis}

We need to implement ``causal interventions'' to validate the hypotheses. Following the approach of \citet{reddy2023mechanistic}, we propose studying a \textit{three-parameter dynamical system}, which simulates only the dynamics of the sequential query mechanism, removing the random guessing mechanism.

\paragraph{Comparing the training dynamics of the three-parameter model with the training dynamics of transformers to validate the causal hypotheses}
Since the random guessing mechanism is removed and the sequential query mechanism is kept, we anticipate that
\begin{enumerate}
    \item Hypothesis~\ref{hyp:random-guess} holds if the three-parameter model loses the slow learning phase in the training dynamics.
    \item Hypothesis~\ref{hyp:sequential-query} holds if the three-parameter model preserves the abrupt phase transition in the training dynamics.
\end{enumerate}

\paragraph{Approximate SoftMax operator and notations for content and buffer spaces.} For convenience, we define the approximate SoftMax operator as:
\[
\pseudosoftmax(u, M) = \frac{\exp(u)}{\exp(u)+M}.
\]
Intuitively, the approximate SoftMax gives the probability of an item with logit $u$ where the remaining $M$ logits are all zero. 
Given a residual state $u$, we use $\content(u)$, $\buffer_1(u)$, $\buffer_2(u)$ to denote the original content (i.e., the token embedding and the positional embedding) of token $u$, the buffer space of $u$ in the first layer, and the buffer space of $u$ in the second layer, respectively. Since the tokens are not semantically related, we assume that $\langle\content(a),\content(b)\rangle = \mathbbm{1}\{a=b\}$ for any two tokens $a$ and $b$, which means that $\{\content(\cdot)\}$ is an orthonormal basis. 

\paragraph{The meaning of three parameters.} The sequential query mechanism consists of a copy layer in the first layer, \text{query}~to \brg~in the second layer, and \text{query}~to \text{\Ed}~in the third layer.
We use parameters $\alpha$, $\beta$, and $\gamma$ to represent the progressive measure of their functionalities.
\begin{align}
&\buffer_1(\text{\targetbridge})=w_1\content(\text{\targetsource}), \label{eqn:L1_copy_brg}\\
&\buffer_1(\text{\targetend})=w_1\content(\text{\targetbridge}), \label{eqn:L1_copy_end}\\
&\buffer_2(\text{query})=w_2\content(\text{\targetbridge}),  \label{eqn:L2_seq_qry}\\
&\texttt{Output}=w_3 \content(\text{\targetend}), \label{eqn:L3_output}\\
&\text{Loss} = -\log[\pseudosoftmax(\xi w_3, V)].\label{eqn:loss}
\end{align}
where
\begin{align*}
w_1&=\pseudosoftmax(\alpha,N), \\
w_2&=\pseudosoftmax(\beta\langle\content(\text{query}),\buffer_1(\text{\targetbridge})\rangle,2N),\\
w_3&=\pseudosoftmax(\gamma\langle\buffer_2(\text{query}),\buffer_1(\text{\targetend})\rangle,2N).
\end{align*}
Note that when we set $\alpha\to\infty$, $\beta\to\infty$, and $\gamma\to\infty$, $\text{Loss}\to 0$, corresponding to the three-layer transformer trained after $10000$ steps. When we set $\alpha=\beta=\gamma=0$, the loss is close to a uniform guess in the vocabulary, corresponding to an untrained three-layer transformer. 

\paragraph{The derivation of Equations~\eqref{eqn:L1_copy_brg} and \eqref{eqn:L1_copy_end}.} We present how we simplify a full transformer block to get Equations~\eqref{eqn:L1_copy_brg} and \eqref{eqn:L1_copy_end}.
As illustrated in Section~\ref{sec:tf-twohop}, the first attention block relies on the positional information to copy parent tokens to the buffer spaces of child tokens. The attention logits are given by
\begin{align*}
~&\attlogit(\child\to\parent)\\
=~&\text{Pos}_i^\top Q^{(1)\top}K^{(1)} \text{Pos}_{i-1}, 
\end{align*}
where $Q^{(1)}$, $K^{(1)}$ are weight matrices in the first layer. We assume that $\text{Pos}_i^\top Q^{(1)\top}K^{(1)} \text{Pos}_{i-1}=\alpha$ for any $i$. Since we reshuffle the positions for $\text{\Brg[1]}$ and $\text{\Ed}$ for each sequence, following \citet{reddy2023mechanistic}, we approximate the attention weights to parent tokens by $\pseudosoftmax(\alpha, N)$, where $N$ comes from taking the average from  $2N$ positions. This gives Equations~\eqref{eqn:L1_copy_brg} and \eqref{eqn:L1_copy_end}.


\paragraph{The derivation of Equation~\eqref{eqn:L2_seq_qry}.} 
Similarly, the $\buffer_2(\text{query})$ is proportional to the attention from the \text{query}~token to $\text{\targetbridge}$ in the second layer. The \text{query}~token uses its $\content(\text{query})$ to fit the $\buffer_1(\text{\targetbridge})$, copying $\content(\text{\targetbridge})$ to the residual stream. Therefore, 
\begin{align*} 
~&\attlogit(\text{query}\to\text{\targetbridge})\\
=~&\content(\text{query})^\top Q^{(2)\top} K^{(2)}\buffer_1(\text{\targetbridge})\\
=~&\beta \cdot \langle \buffer_1(\text{\targetbridge}), \content(\text{query}) \rangle,
\end{align*}
where the last line could be viewed as a re-parametrization of $Q^{(2)\top} K^{(2)}$, with $\beta \propto \|Q^{(2)\top} K^{(2)}\|_2$. Moreover, we fix the attention logits from $\text{query}$ token to all other tokens to be zero, removing  mechanisms other than the sequential query. The attention weight from the \text{query}~token to the \text{\targetbridge}~becomes $\pseudosoftmax(\attlogit(\text{query}\to\text{\targetbridge}), 2N)$. This gives Equation~\eqref{eqn:L2_seq_qry}.

\paragraph{The derivation of Equation~\eqref{eqn:L3_output} and \eqref{eqn:loss}.}
The $\text{query}$ token increasingly concentrates on the \text{\targetend}~token along the training dynamics. With the same manner of Equation~\eqref{eqn:L2_seq_qry}, we set that
\begin{align*}
~& \attlogit(\text{query}\to\text{\targetend})  \\
=~&\buffer_2(\text{query}) Q^{(3)\top} K^{(3)}\buffer_1(\text{\targetend})\\
= ~& \gamma \cdot \langle \buffer_2(\text{query}), \buffer_1(\text{\targetend})\rangle.
\end{align*}
We focus on  $\attlogit(\text{query}\to\text{\targetend})$ and set all other \attlogit~to be zero.
The attention weight from the query to the \text{\targetend}~becomes $\pseudosoftmax(\attlogit(\text{query}\to\text{\targetend}), 2N)$. We first consider the output \logit~on the query token. Through the logit lens, as illustrated in \Cref{fig:logit-lens-layer-3}, the value states of \text{\targetend}~tokens have large logits on themselves. Therefore, we assume that $\readout[\Val(\text{\targetend})] = \xi \cdot \bm{e}_{\text{\targetend}} \in \R^V$, with $\xi>0$ and $\bm{e}_{\text{\targetend}}$ being a one-hot vector in $\R^V$ that is non-zero on the index of $\text{\targetend}$. In our simulation, we fix $\xi=30$.
The loss can therefore be approximated through Equation~\eqref{eqn:loss}.

\begin{figure}[h]
    \centering
    \begin{subfigure}[t]{0.23\textwidth}
        \caption{\small 3-layer transformer}
        \includegraphics[width=\textwidth]{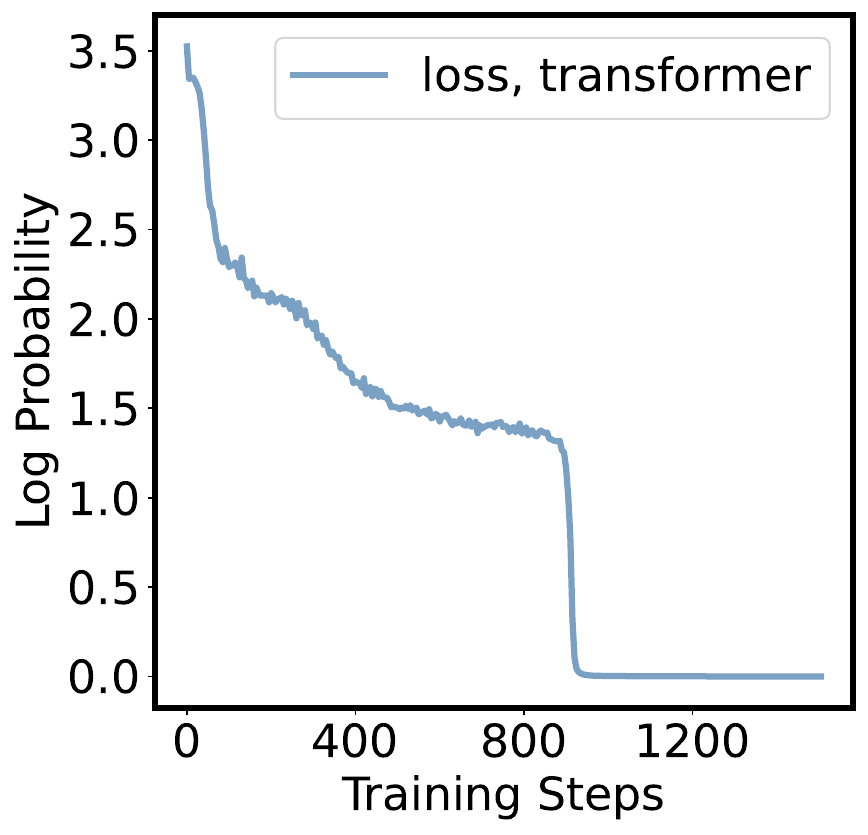}
    \end{subfigure}
    \begin{subfigure}[t]{0.23\textwidth}
        \centering 
        \caption{\small 3-parameter model}
\includegraphics[width=\textwidth]{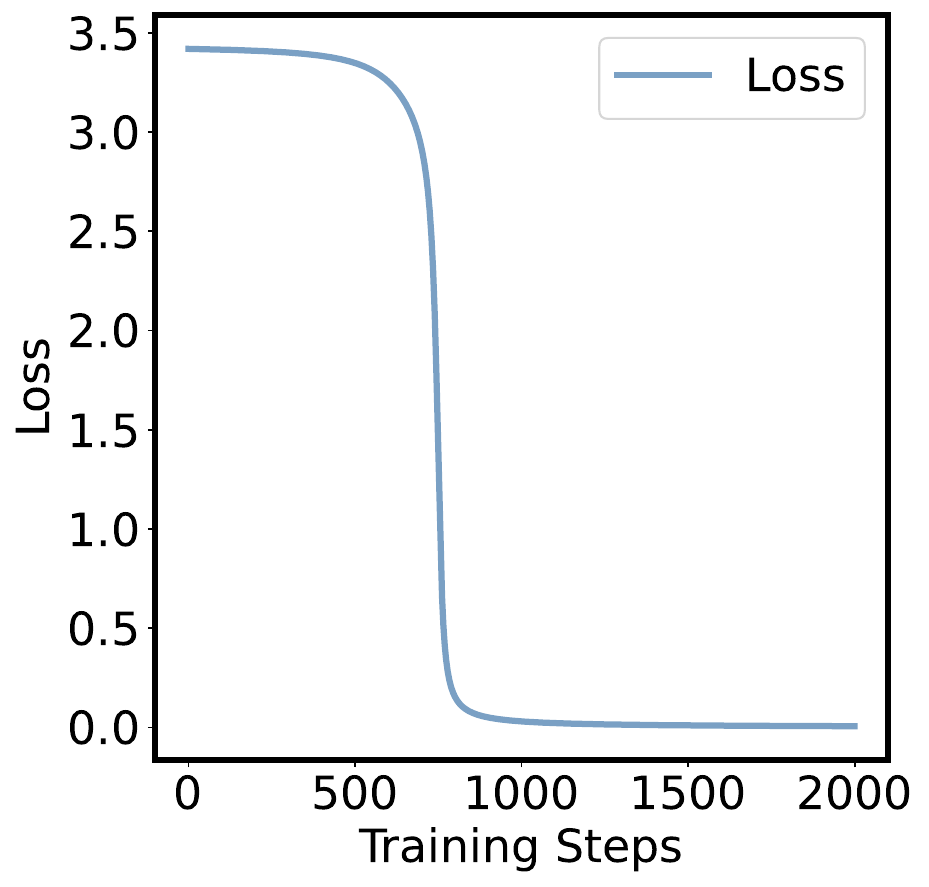}
    \end{subfigure}
    \begin{subfigure}[t]{0.23\textwidth}
        \caption{\small 3-layer transformer}
        \includegraphics[width=\textwidth]{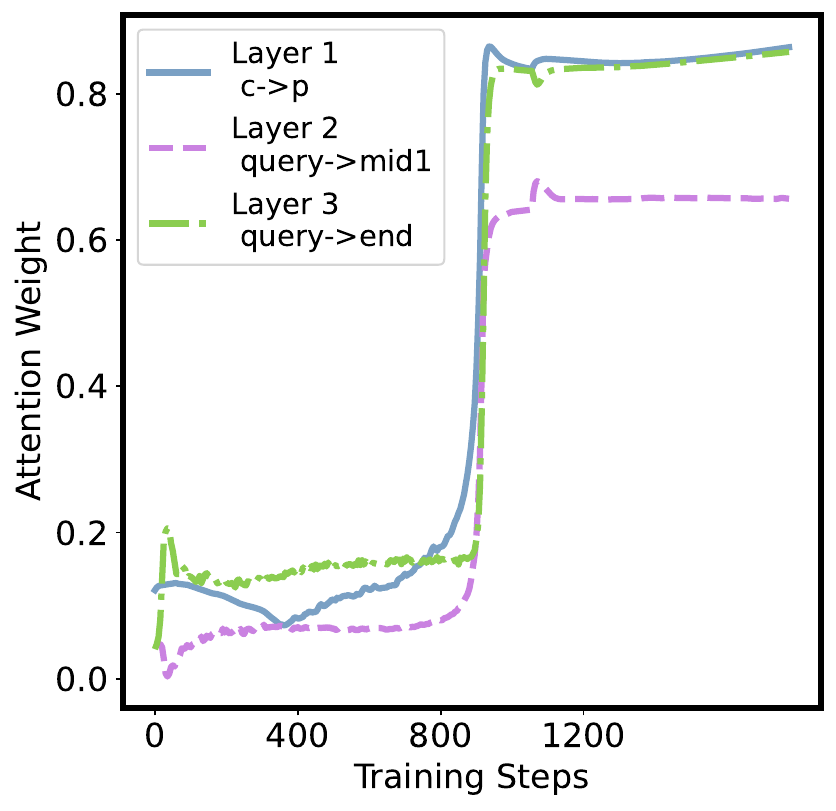}
    \end{subfigure}
    \begin{subfigure}[t]{0.23\textwidth}
    \centering
        \subcaption{\small 3-parameter model}
\includegraphics[width=\textwidth]{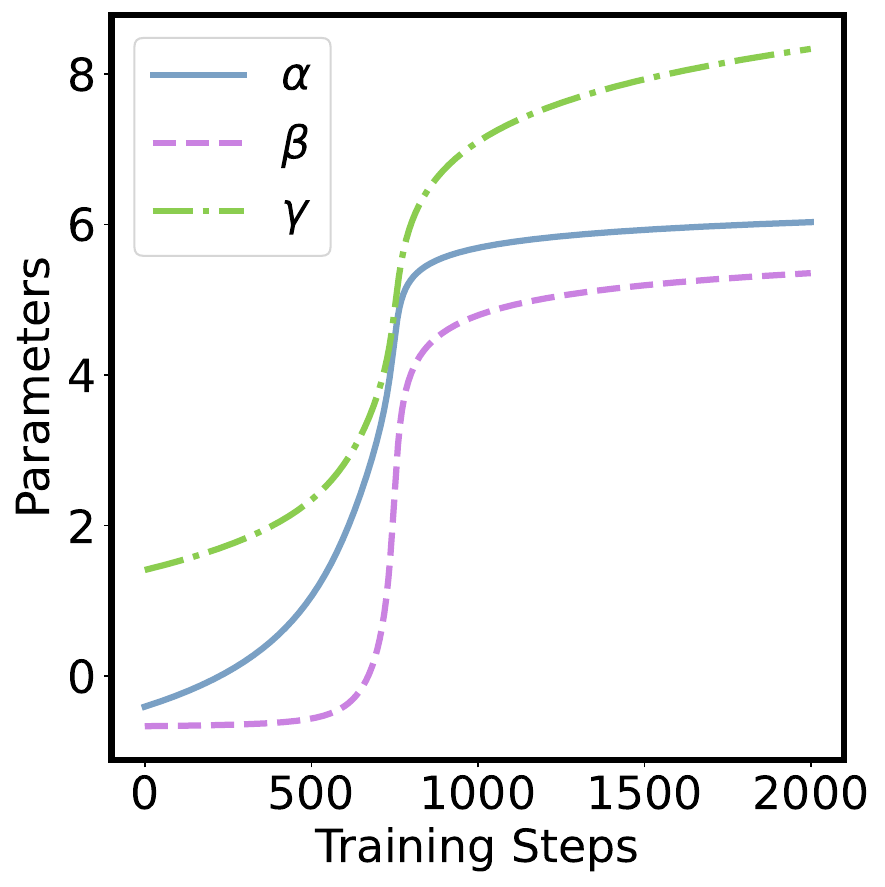}
    \end{subfigure}
    \caption{\textbf{The comparison between 3-layer transformer and 3-parameter model validates our causal hypotheses.} \textit{Top left (a)}: The loss dynamics of the 3-layer transformer shows a slow learning phase (0-400 steps) followed by an abrupt phase transition (around 800 steps). \textit{Top right (b)}: The 3-parameter model, which only simulates the sequential query mechanism, skips the slow learning phase but preserves the abrupt phase transition, validating both hypotheses. \textit{Bottom left (c)}: The dynamics of important attention weights for the sequential query mechanism. \textit{Bottom right (d)}: The parameter dynamics of the 3-parameter model shows synchronized phase transitions in all three parameters ($\alpha$, $\beta$, $\gamma$), indicating the formation of the sequential query mechanism.}

    \label{fig:3_param_dnamics}
\end{figure}

\paragraph{Simulations on three-parameter model validate Hypotheses~\ref{hyp:random-guess} and~\ref{hyp:sequential-query}.} We optimize the loss function in~\cref{eqn:loss} by gradient descent with learning rate $0.1$.  
\Cref{fig:3_param_dnamics} presents the training dynamics of the 3-layer transoformer and the 3-parameter model. Since the model does not incorporate the random guessing mechanism, the loss remains unchanged during the first $1000$ steps, \hanlin{For the 3-param model, it's not 1000 epochs.} validating Hypothesis~\ref{hyp:random-guess}. 
Both the parameters and the loss function go through a sudden phase transition around step $1000$, suggesting that the emergence of the sequential query mechanism is the driving force behind the abrupt drop in loss. This validates Hypothesis~\ref{hyp:sequential-query}.

\end{document}